\documentclass[lettersize,journal]{IEEEtran}

\usepackage{amsmath,amsfonts,amssymb}
\usepackage[inline]{enumitem}
\usepackage{textcomp}
\usepackage{stfloats}
\usepackage{url}
\usepackage{verbatim}
\usepackage{graphicx}
\usepackage{multirow}
\usepackage{subfigure}
\usepackage{float}
\usepackage{comment}
\usepackage{bbm}
\newcommand{\etal}{\textit{et al.}}

\usepackage[backend=bibtex,style=numeric,natbib=true,citestyle=numeric,sorting=none,backref=true,backrefstyle=three,maxcitenames=2]{biblatex}
\bibliography{refs}
\usepackage[breaklinks=true,colorlinks,bookmarks=false]{hyperref}

\usepackage[capitalise,noabbrev]{cleveref}
\usepackage[pagewise]{lineno}

\newcount\Comments  
\usepackage{color}
\definecolor{darkgreen}{rgb}{0,0.5,0}
\definecolor{purple}{rgb}{1,0,1}
\newcommand{\kibitz}[2]{\ifnum\Comments=1\textcolor{#1}{#2}\fi}
\newcommand{\nassim}[1]{\kibitz{darkgreen}      {[Nassim: #1]}}
\newcommand{\yi}[1]  {\kibitz{blue}   {[Yi: #1]}}

\usepackage[
monochrome
]{xcolor}

\usepackage{fancyhdr}
\fancypagestyle{copyright}{%
    \fancyhf{}
    \cfoot{\footnotesize%
        \textcopyright~2022 IEEE. Personal use of this material is permitted. Permission from IEEE must be obtained for all other uses, in any current or future media, including reprinting\slash republishing this material for advertising or promotional purposes, creating new collective works, for resale or redistribution to servers or lists, or reuse of any copyrighted component of this work in other works.    
    }
}

\begin{document}
\title{Self-supervised Learning in Remote Sensing: A Review}

\author{Yi Wang,~\IEEEmembership{Student Member,~IEEE,}
Conrad M Albrecht,~\IEEEmembership{Member,~IEEE,}
Nassim Ait Ali Braham, Lichao Mou,
Xiao Xiang Zhu,~\IEEEmembership{Fellow,~IEEE}

\thanks{The work is jointly supported by the Helmholtz Association
through the Framework of Helmholtz AI (grant  number:  ZT-I-PF-5-01) - Local Unit ``Munich Unit @Aeronautics, Space and Transport (MASTr)'' and Helmholtz Excellent Professorship ``Data Science in Earth Observation - Big Data Fusion for Urban Research''(grant number: W2-W3-100), by the German Federal Ministry of Education and Research (BMBF) in the framework of the international future AI lab "AI4EO -- Artificial Intelligence for Earth Observation: Reasoning, Uncertainties, Ethics and Beyond" (grant number: 01DD20001) and by German Federal Ministry for Economic Affairs and Climate Action in the framework of the "national center of excellence ML4Earth" (grant number: 50EE2201C). \\
\textit{(Corresponding author: Xiao Xiang Zhu.)}}
\thanks{Y. Wang, N.A.A. Braham, L. Mou and X. Zhu are with the Remote Sensing Technology Institute (IMF), German Aerospace Center (DLR), Germany and with the Chair of Data Science in Earth Observation, Technical University of Munich (TUM), Germany (e-mails: yi.wang@dlr.de; Nassim.aitalibraham@dlr.de; lichao.mou@dlr.de; xiaoxiang.zhu@dlr.de).}
\thanks{C. M Albrecht is with the Remote Sensing Technology Institute (IMF), German Aerospace Center (DLR), Germany (e-mail: conrad.albrecht@dlr.de).}

}

\markboth{ACCEPTED BY IEEE GEOSCIENCE AND REMOTE SENSING MAGAZINE, 2022}%
{Shell \MakeLowercase{\textit{et al.}}: A Sample Article Using IEEEtran.cls for IEEE Journals}


\maketitle
\begin{abstract}
In deep learning research, self-supervised learning (SSL) has received great attention triggering interest within both the computer vision and remote sensing communities. While there has been a big success in computer vision, most of the potential of SSL in the domain of earth observation remains locked. In this paper, we provide an introduction to, and a review of the concepts and latest developments in SSL for computer vision in the context of remote sensing. Further, we provide a preliminary benchmark of modern SSL algorithms on popular remote sensing datasets, verifying the potential of SSL in remote sensing and providing an extended study on data augmentations. Finally, we identify a list of promising directions of future research in SSL for earth observation (SSL4EO) to pave the way for fruitful interaction of both domains.
\end{abstract}

\begin{IEEEkeywords}
Deep learning, self-supervised learning, computer vision, remote sensing, Earth observation.
\end{IEEEkeywords}

\thispagestyle{copyright}

\section{Introduction}

\IEEEPARstart{A}{dvances} of deep neural networks to model the rich structure of large amounts of data has led to major breakthroughs in computer vision, natural language processing, automatic speech recognition, and time series analysis~\cite{lecun2015deep}. However, the performance of deep neural networks is very sensitive to the size and quality of the training data. Thus, a plurality of annotated datasets (e.g. ImageNet~\cite{deng2009imagenet}) have been generated for supervised training in the last decade,
\nassim{TODO: Add a couple of references, e.g. Imagenet}\yi{ImageNet added}driving progress in many fields. Unfortunately, annotating large-scale datasets is an extremely laborious, time-consuming, and expensive procedure. This limitation of the supervised learning paradigm strongly impedes the applicability of deep learning in real-world scenarios. 

\begin{figure}
\centering
\includegraphics[width=0.9\linewidth]{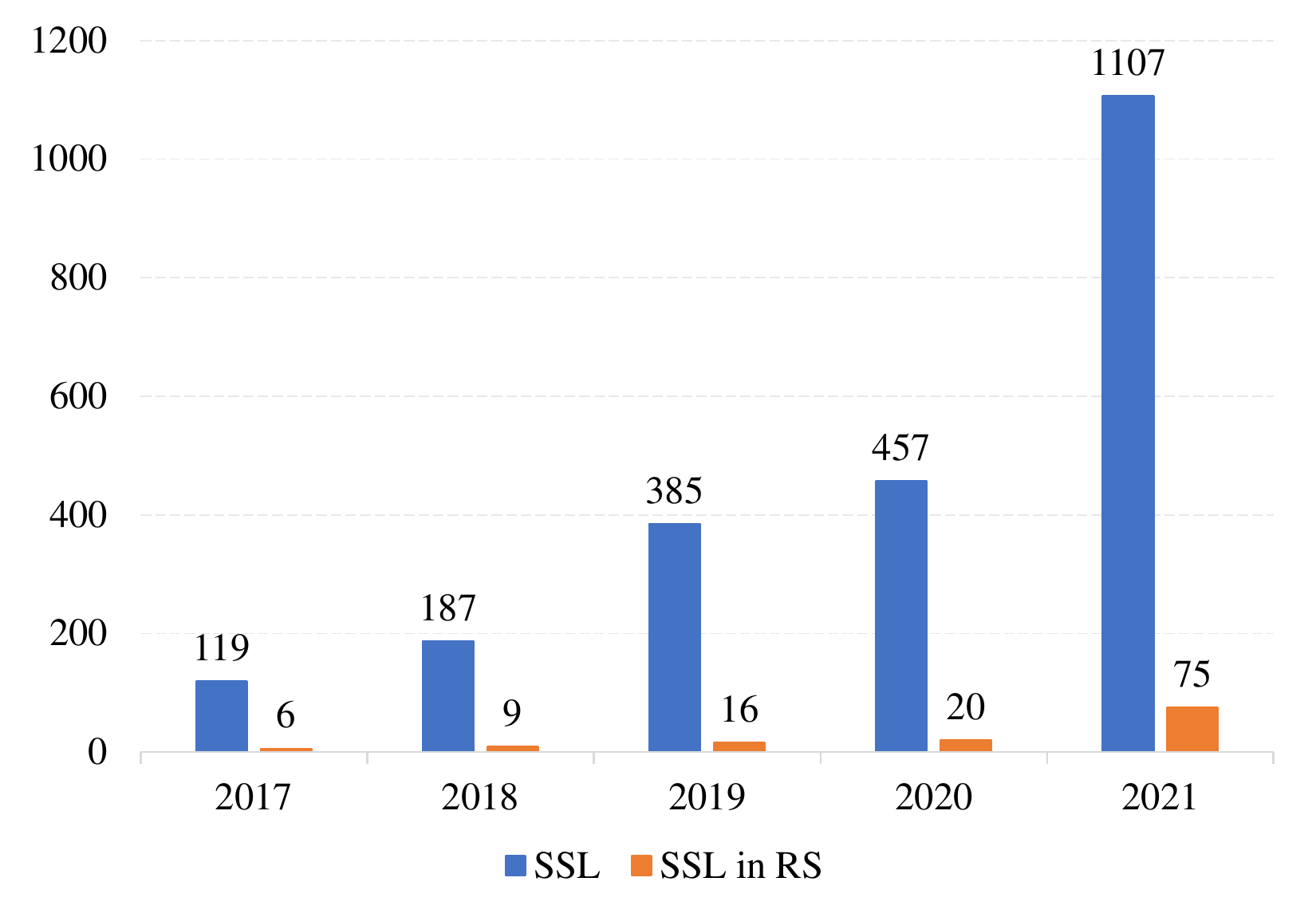}
\caption[number-of-publications]{
\textcolor{blue}{The number of recent publications related to self-supervised learning (SSL). While a clear trend of increased efforts to advance SSL is observed, activity in remote sensing lags behind.}}
\label{fig:number-of-publications}
\end{figure}

A lot of research efforts have been deployed to tackle the challenge of data annotation in the machine learning literature. In fact, numerous alternatives to vanilla supervised learning have been studied, such as \nassim{See if changing this later, otherwise I will add refs}\yi{added}
unsupervised learning~\cite{dike2018unsupervised}, semi-supervised learning~\cite{van2020survey}, weakly supervised learning~\cite{zhou2018brief} and meta-learning~\cite{vanschoren2019meta}. Recently, \textit{self-supervised learning (SSL)} did raise considerable attention in computer vision (see Fig. \ref{fig:number-of-publications}) and achieved significant milestones towards the reduction of human supervision. Indeed, by distilling representative features from unlabeled data, SSL algorithms are already outperforming supervised pre-training on many problems~\cite{goyal2021self}.
\nassim{I should add a ref}\yi{added}

Meanwhile, the success of deep learning did also spark remarkable progress in remote sensing~\cite{zhu2017deep}. Yet, data annotation remains a major challenge for earth observation as well. In that regard, self-supervised learning is a promising avenue of research for remote sensing. While there exist comprehensive surveys of SSL in computer vision~\cite{jing2020self,jaiswal2021survey,liu2021self,ohri2021review}, a didactic and up-to-date introduction to the remote sensing community is missing. This review aims at bridging this gap, documenting progress in self-supervised visual representation learning, and how it did get applied in remote sensing to date. Specifically:
\begin{itemize}
    \item To the remote sensing community, we provide a comprehensive introduction and literature review on self-supervised visual representation learning.
    \item We summarize a taxonomy of self-supervised methods covering existing works from both computer vision and remote sensing communities.
    \item We provide quantification of performance benchmarks on established SSL methodologies utilizing the \textit{ImageNet} dataset~\cite{deng2009imagenet}, and extend the analysis to three multi-spectral satellite imagery datasets: \textit{BigEarthNet}~\cite{sumbul2019bigearthnet}, \textit{SEN12MS}~\cite{schmitt2019sen12ms}, 
    and \textit{So2Sat-LCZ42}~\cite{zhu2019so2sat}.
    \item We discuss the link of natural imagery and remotely sensed data, providing insights for future work in self-supervised learning for remote sensing and earth observation.
\end{itemize}

\subsection{Supervision: Data vs.\ Labels}

A significant fraction of real-world applications of machine\slash deep learning methodologies calls for massive amounts of training data. Fortunately, the increasing adoption of open access in earth observation provides remote sensing researchers with a plethora of such. However, annotation of such vast amounts of earth observation data implies significant human interaction with frequent updates needed every single day. For example, apart from the large amounts of sampled data to label, a single large-scale dataset may require separate annotation depending on downstream applications. Moreover, datasets sensing the physical world get outdated over time due to both natural and man-made changes. Both aspects significantly amplify the efforts in annotating remotely sensed data of the Earth.

Another big challenge of machine learning in remote sensing is label noise. Due to the quality of remotely sensed images and the complexity of various specific applications, it is very difficult to generate perfect labels during large-scale data annotation. Therefore, there always exists a trade-off between the number of annotations and their quality: large but noisy labeled datasets can bias the model, whereas small amounts of good-quality labels usually lead to overfitting.

In addition, it has become popular to execute \textit{pre-training} on established benchmark datasets before \textit{fine-tuning} deeply learned models for downstream tasks that do not have enough labels. This procedure is commonly referred to as transfer learning~\cite{zhuang2020comprehensive}. However, the performance of supervised pre--training \textcolor{blue}{depends} on the domain difference between source and target data. A good pre--trained model renders well on a similar dataset but is not as useful on a very different one. 

All the above challenges emphasize the growing gap between an increasing amount of remote sensing data and the shortage of good-quality labels, calling for techniques to exploit the corpus of unlabeled data to learn valuable information that easily transfers to multiple applications. Self-supervised learning offers a paradigm to approach this dilemma.

\subsection{Self-supervision: Learning Task-agnostic Representations of Data}

\begin{figure}
    \centering
    \includegraphics[width=.9\linewidth]{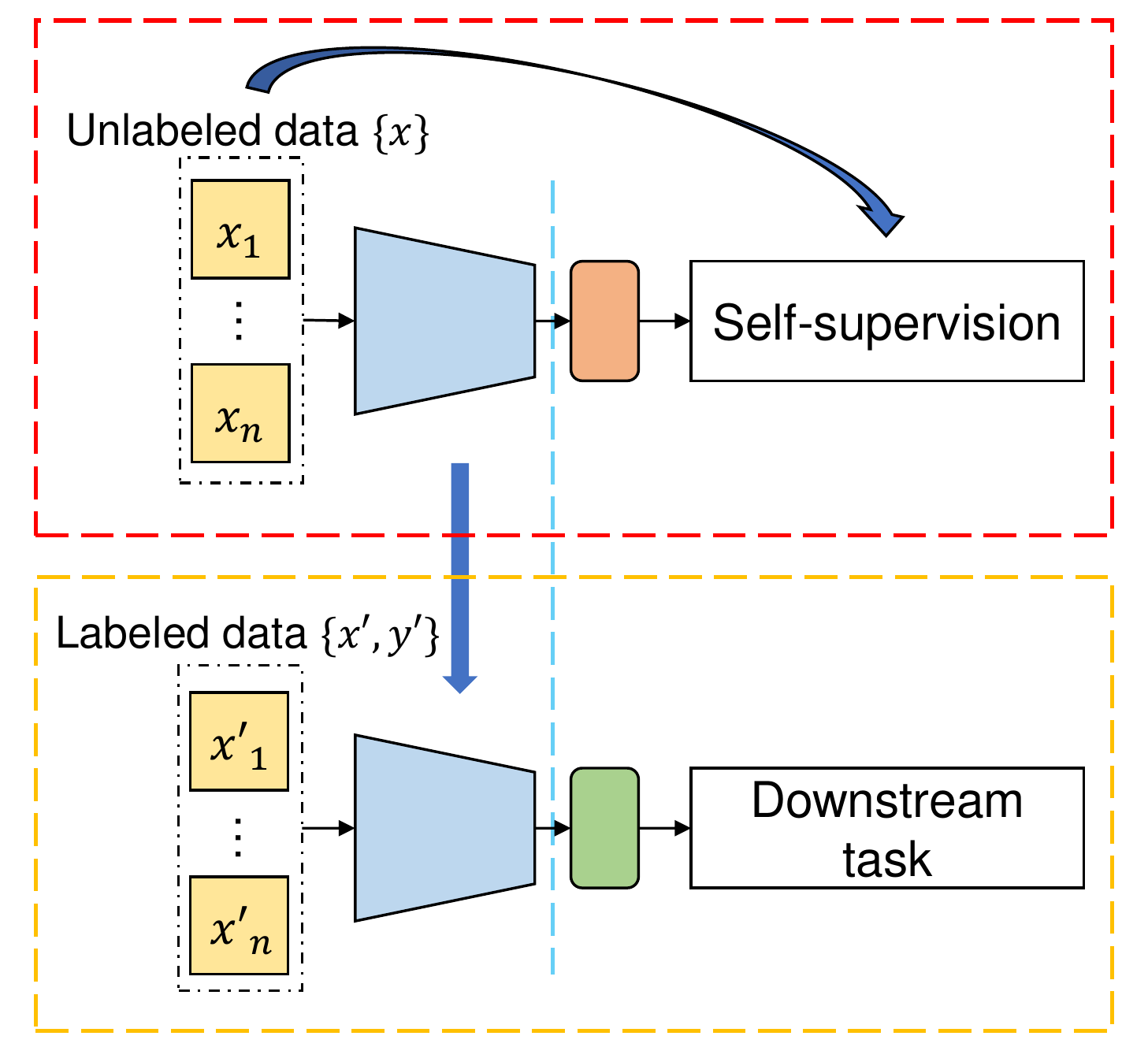}
    \caption{The general pipeline of self-supervised learning. The visual representation is learned through self-supervision that comes from the unlabeled data. The learned parameters serve as a pre-trained model and are transferred to supervised downstream tasks for fine-tuning.}
    \label{fig:SelfSupervisionScheme}
\end{figure}

Fig. \ref{fig:SelfSupervisionScheme} schematically depicts the general underlying principle of self-supervised learning. Based on a certain self-produced objective (the so-called \textit{self-supervision}), a large amount of unlabeled data is exploited to train a model $f$\footnote{Typically a convolutional neural network (CNN), e.g. ResNet~\cite{he2016deep}. Recent work also show promising results with vision transformers (ViT)~\cite{dosovitskiy2020image}.} by optimizing this objective without requiring any manual annotation. With a carefully designed self-supervised problem, the model $f$ gets the ability to capture high-level representations of the input data. Afterward, the model $f$ can be further transferred to supervised downstream tasks for real-world applications.  

The most common strategies to design such self-supervision typically exploit three types of objectives: $(1)$ reconstructing input data $x$, $f(x)\rightarrow x$, $(2)$ predicting a self-produced label $c$ which usually comes from contextual information and data augmentation (e.g., predicting the rotation angle of a rotated image), $f(x)\rightarrow c$, $(3)$ contrasting semantically similar inputs $x_1$ and $x_2$ (e.g., the encoded features of two augmented views of the same image should be identical), $|f(x_1)-f(x_2)|\rightarrow 0$.

Given self-supervised training succeeded, the pre-trained model $f$ may get transferred to downstream tasks. As Opposed to supervised pre-training, models pre-trained by self-supervision bear the potential to leverage more general representations and offer a paradigm to mitigate the shortcomings of supervised learning: $(1)$ no human annotation is needed for pre-training, $(2)$ small amounts of labels are sufficient for good performance on downstream tasks, $(3)$ \textcolor{blue}{little domain gap between pre-training and downstream dataset can be ensured by collecting unlabeled data from the target application.}

\textbf{Link to semi-supervised learning.} Self-supervised learning algorithms may be considered as part of semi-supervised learning---a branch of machine learning concerned with labeled and unlabelled data. Van Engelen \etal~\cite{van2020survey} proposed a comprehensive taxonomy of semi-supervised classification algorithms based on conceptual and methodological aspects for processing unlabelled data. Within that scheme, self-supervised learning may get categorized as \textit{unsupervised preprocessing}: in the first step, unlabeled data gets transformed to extract feature representations; labeled data is then utilized to adapt the model to specific tasks.

Within the remote sensing community, semi-supervised learning has been long studied and enjoys applications in, e.g., hyperspectral image recognition and processing~\cite{dopido2013semisupervised,ratle2010semisupervised,wang2018self,hong2019wu,hong2019learning,hong2020joint,cao2020sdfl,cao20203d,li2020deep,yue2021self}, multi-spectral image segmentation~\cite{hong2019cospace,hong2019learnable,hong2020x,hu2020unsupervised,hua2021semantic,hua2021aerial,saha2020semisupervised} and SAR-optical data fusion~\cite{hu2019mima}.

\textbf{Link to unsupervised learning.} Self-supervised learning is also a discipline in the realm of unsupervised learning---the general set of machine learning methods that are independent of human annotation. \textcolor{blue}{However, it is very ambiguous in various communities the difference between self-supervised and unsupervised learning. In this paper, we separate them in terms of methodology. Traditional unsupervised algorithms tend to utilize statistics of the input data and generate groups by dimensionality reduction~\cite{abdi2010principal,cheriyadat2013unsupervised} or clustering~\cite{jain1999data}. On the other hand, self-supervised learning is a recent terminology that refers to approaches in which a model (e.g., a neural network) is trained to learn good data representations using supervision signals that are automatically generated from the data itself.}

In remote sensing, unsupervised learning has also been actively employed in various applications, e.g., in scene classification~\cite{li2016unsupervisedrs,cheriyadat2013unsupervised,romero2015unsupervised}, semantic segmentation~\cite{bandyopadhyay2007multiobjective,fan2009single,saha2020unsupervised}, change detection~\cite{ghosh2011fuzzy,munyati2004use} and multi-sensor data analysis~\cite{zabalza2014novel,cao2003remote}.

\subsection{Performance Evaluation in Self-Supervised Learning}

To evaluate the performance of self-supervised learning methods, downstream tasks are commonly defined in order to transfer the pre-trained model $f$ to specific applications, such as scene classification, semantic segmentation, and object detection. The performance of transfer learning on these high-level vision tasks estimates the generalizability of the model $f$. In practice, three common procedures are used for quantitative evaluation: 
\begin{itemize}
    \item \textbf{Linear probing (or linear classification)}: refers to fixing the parameters of the learned model $f$ (frozen encoder), and training a linear classifier $g$ on top of the generated representations. This approach measures how linearly separable the embeddings produced by the pre-trained model $f$ are.
    \item \textbf{$K$ nearest neighbors (KNN)}: refers to applying weighted voting of the $K$ nearest neighbors of the input test image $x$ in the feature space. This approach is non-parametric and can be used for classification tasks. 
    \item \textbf{Fine-tuning}: refers to training a model on the downstream task by using the parameters of the pre-trained model $f$ as an initialization. It is the most general procedure since it is not only limited to classification.  
\end{itemize}

In addition to quantitative evaluation, qualitative visualization methods can also be used to derive insights on the features learnt during the self-supervised training of the encoder $f$. Among these, two popular techniques are kernel visualization~\cite{gidaris2018unsupervised,noroozi2018boosting,caron2018deep} for CNN and feature map visualization~\cite{gidaris2018unsupervised,caron2021emerging,el2021xcit} for both CNN and ViT. Kernel visualization plots kernels of the first convolutional layer of $f$ and compares it with kernels from corresponding fully supervised training. The similarity of kernels learned by supervised and self-supervised training sheds light on the efficacy of self-supervision. Along the lines, feature map visualization displays hidden layer feature maps of $f$ in order to analyze the spatial attention of $f$ on the input $x$. Once \textcolor{blue}{again}, the feature maps are compared with their counterparts of supervised training for reference.

\section{Generative, Predictive, and Contrastive: A taxonomy of self-supervised learning}

\begin{figure}
\centering
\includegraphics[width=.9\linewidth]{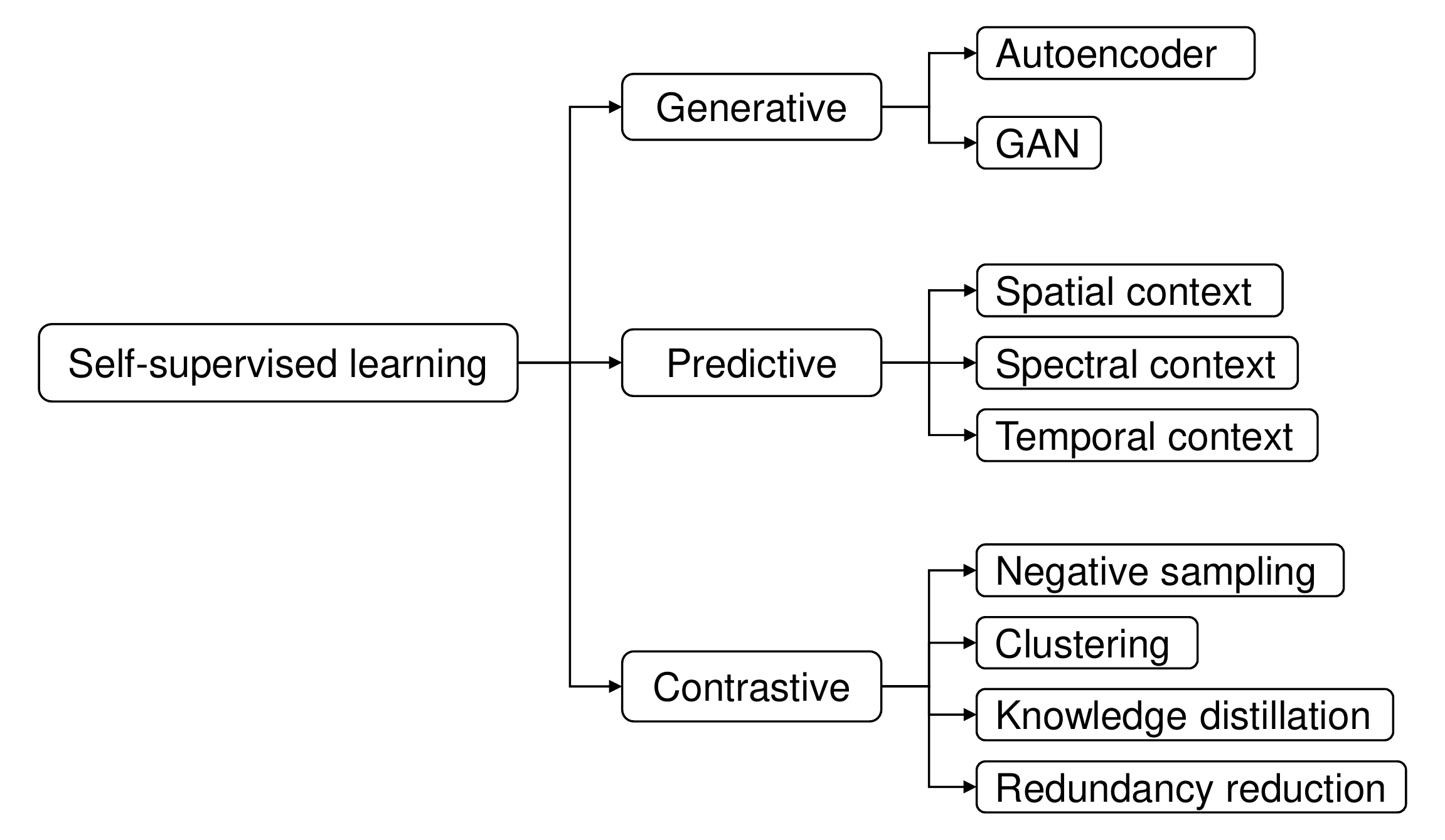}
\caption[taxonomy-ssl]{A taxonomy of self-supervised learning.}
\label{fig:taxonomy-ssl}
\end{figure}

In this section, we review works in self-supervised learning following an extended
taxonomy: \textit{generative}, \textit{contrastive}, and \textit{predictive} methods, as displayed in Fig. \ref{fig:taxonomy-ssl}. Generative methods learn to reconstruct or generate input data, predictive methods learn to predict self-generated labels, and contrastive methods learn to maximize the similarity between semantically identical inputs. While some existing categorizations solely classify self-supervised methods into generative and contrastive~\cite{jaiswal2021survey,liu2021self}, we add the \textit{predictive} category conceptually based on the processing level of supervision (i.e., the self-supervision for generative methods is raw input, while for predictive methods it is carefully designed labels with high-level semantic information that come after processing of the input).\nassim{What does it mean?} \yi{modified}Adopting such categorization also provides a historical perspective to the development of self-supervised learning. In the following sub-sections, we introduce in detail the three types of self-supervised methodologies in computer vision and link them to works in remote sensing for a side-by-side comparison.

\begin{table*}
\caption{a gallery of representative self-supervised methods.}
\centering

\begin{tabular}{|c|c|l|}
\hline
Category                                          & Sub-category                            & \multicolumn{1}{c|}{Representative methods}                                                                       \\ \hline
\multicolumn{1}{|l|}{\multirow{8}{*}{Generative}} & \multirow{5}{*}{Autoencoder}            & Autoencoder (AE)~\cite{ballard1987modular}: encode and reconstruct the input.                                     \\
\multicolumn{1}{|l|}{}                            &                                         & Sparse AE~\cite{ng2011sparse}: sparsity constraints on the hidden units.                                          \\
\multicolumn{1}{|l|}{}                            &                                         & Denoising AE~\cite{vincent2010stacked}: reconstruct clear image from noisy input.                                 \\
\multicolumn{1}{|l|}{}                            &                                         & VAE~\cite{kingma2013auto}: encode the input to a normal distribution.                                             \\
\multicolumn{1}{|l|}{}                            &                                         & MAE~\cite{he2021masked}: reconstruct randomly masked patches with vision transformer.                             \\ \cline{2-3} 
\multicolumn{1}{|l|}{}                            & \multirow{3}{*}{GAN}                    & GAN~\cite{goodfellow2014generative}: adversarial training with a generator and a discriminator.                   \\
\multicolumn{1}{|l|}{}                            &                                         & AAE~\cite{makhzani2015adversarial}: adversarial variational autoencoder.                                          \\
\multicolumn{1}{|l|}{}                            &                                         & BiGAN~\cite{donahue2016adversarial}: an additional encoder to map samples to representations.                     \\ \hline
\multirow{9}{*}{Predictive}                       & \multirow{4}{*}{Spatial}                & Relative position~\cite{doersch2015unsupervised}: predict relative position of random patch pairs.                \\
                                                  &                                         & Jigsaw~\cite{noroozi2016unsupervised}: predict the correct order of a jigsaw puzzle.                              \\
                                                  &                                         & Rotation~\cite{gidaris2018unsupervised}: predict the rotation angle of rotated images.                            \\
                                                  &                                         & Inpainting~\cite{pathak2016context}: recover a missing patch of the input image.                                  \\ \cline{2-3} 
                                                  & Spectral                                & Colorization~\cite{zhang2016colorful}: predict colorful image from gray-scale input.                              \\ \cline{2-3} 
                                                  & Temporal                                & Frame order~\cite{kim2019self}: predict the order of frame sequences.                                             \\ \cline{2-3} 
                                                  & \multirow{3}{*}{Others}                 & Counting~\cite{noroozi2017representation}: count the visual primitives within the input image.                    \\
                                                  &                                         & Artifact~\cite{jenni2018self}: spot and predict artifacts.                                                        \\
                                                  &                                         & Audio~\cite{owens2016ambient}: predict a statistical summary of the sound associated with a video.                \\ \hline
\multirow{20}{*}{Contrastive}                     & \multirow{8}{*}{Negative sampling}      & Triplet loss~\cite{chopra2005learning}: learn a similarity metric discriminatively with triplet loss.             \\
                                                  &                                         & CPC~\cite{oord2018representation}: contrastive predictive coding with InfoNCE loss.                               \\
                                                  &                                         & DIM~\cite{hjelm2018learning}: maximize mututal information between global-local feature pairs.                    \\
                                                  &                                         & InstDisc~\cite{wu2018unsupervised}: maximally scatter the features of different images over a unit sphere.        \\
                                                  &                                         & PIRL~\cite{misra2020self}: pretext tasks as data augmentation.                                                     \\
                                                  &                                         & MoCo~\cite{he2020momentum}: store negative samples in a queue and momentum-update the key encoder.                \\
                                                  &                                         & SimCLR~\cite{chen2020simple}: end-to-end contrastive structure with large batchsize and strong data augmentation. \\
                                                  &                                         & MoCo-v3~\cite{chen2021empirical}: introduce vision transformer as the encoder backbone for MoCo-v2~\cite{chen2020improved}.               \\ \cline{2-3} 
                                                  & \multirow{5}{*}{Clustering}             & DeepCluster~\cite{caron2018deep}: iteratively leverage k-means clustering to yield pseudo labels for prediction.  \\
                                                  &                                         & LocalAgg~\cite{zhuang2019local}: optimize a local soft-clustering metric.                                         \\
                                                  &                                         & SeLa~\cite{asano2019self}: solve an optimal transport problem to obtain the pseudo-labels.                        \\
                                                  &                                         & SwAV~\cite{caron2020unsupervised}: contrastive learning with online clustering.                                   \\
                                                  &                                         & PCL~\cite{li2020prototypical}: perform iterative clustering and representation learning in an EM-based framework. \\ \cline{2-3} 
                                                  & \multirow{5}{*}{Knowledge distillation} & BYOL~\cite{grill2020bootstrap}: mean-teacher network with a predictor on top of the teacher encoder.              \\
                                                  &                                         & SimSiam~\cite{chen2021exploring}: explore the simplest design of contrastive self-supervised learning.            \\
                                                  &                                         & DINO~\cite{caron2021emerging}: explores the self-distillation scheme with vision transformer backbones.           \\
                                                  &                                         & EsViT~\cite{li2021efficient}: improve DINO with additional region-level contrastive task.                         \\
                                                  &                                         & iBOT~\cite{zhou2021ibot}: cross-view and in-view self-distillation with masked image modeling.                      \\ \cline{2-3} 
                                                  & \multirow{2}{*}{Redundancy reduction}   & Barlow Twins~\cite{zbontar2021barlow}: redundancy reduction on the cross-correlation matrix between features.      \\
                                                  &                                         & VICReg~\cite{bardes2021vicreg}: variance-invariance-covariance regularization.                                    \\ \hline
\end{tabular}

\label{tab:ssl}
\end{table*}

\subsection{Generative Methods}
\label{sec:GenerativeMethods}

\textcolor{blue}{Generative self-supervised methods learn representations by reconstructing or generating input data. A prominent set of methods in this category are autoencoders (AE), which train an encoder $E$ to map input $x$ to a latent vector $z=E(x)$, and a decoder $D$ to reconstruct $x=D(z)$ from $z$.} While $E$ serves the purpose of $f$ (the feature extractor), the joint function $D\circ E$ contributes to a \textit{self-supervised} loss: $\vert\vert x-D(E(x))\vert\vert$.  Another class of methods, generative adversarial networks (GAN), approach the data generation problem from a game theory perspective. In a nutshell, a GAN consists of two models: a generator $\mathcal{G}$ and a discriminator $\mathcal{D}$. The generator takes as input a random vector $z$ and outputs a synthetic sample $\overline{x}=\mathcal{G}(z)$. At the same time, the discriminator is trained to distinguish between real samples $\{x\}$ from the training dataset and synthetic samples generated by $\mathcal{G}$. By doing so, the distribution $P_z$ of synthetic data gradually converges to the distribution $P_z$ of the training dataset, thereby leading to realistic data generation. The vanilla GAN was not designed for feature extraction, but since formally $\mathcal{G}^{-1}=f$, there exist GAN-inspired methods to construct representations $z$.

\subsubsection{Autoencoder (AE)}

The concept of autoencoders was introduced as early as in~\cite{ballard1987modular} to serve as
pre-training of artificial neural networks. Conceptually, an autoencoder $D\circ E$ is a feed-forward $E$ncoder--$D$ecoder network trained to reproduce its input at the output layer. However, one can easily imagine an autoencoder to fail: should $E=D=\mathbbm{1}$, the autoencoder will learn a trivial identity mapping, $D\circ E=\mathbbm{1}$. Thus demanding the latent vector $z$ to retain information on $x$, by itself, is not sufficient to yield expressive representations. Constraints are needed to prevent such scenarios. For example, in early practices, the encoder $E$ typically resembles a bottleneck-like structure such that the dimension of the input data gets compressed down, i.e.\ $\dim x > \dim z$. This links autoencoders to dimensionality reduction. In fact, a loose relation between linear autoencoders and Principal Component Analysis (PCA) \cite{wold1987principal} has previously been studied in the literature~\cite{plaut2018principal}.

Constraining a small dimension of $z$ is not a must to prevent identity mapping. It is also an option to construct $D\circ E$ such that the dimension of $z$ is greater than the input's $x$ dimension with additional sparsity constraints, building so-called \textit{sparse autoencoder}~\cite{ng2011sparse}. Other popular approaches include: \textit{denoising autoencoder}~\cite{vincent2010stacked} and \textit{variational autoencoder} (VAE)~\cite{kingma2013auto}.

Denoising autoencoder is trained to enforce robustness against noise in data. Its reconstruction loss
gets modified to $\vert\vert x - D(E(x+\epsilon)) \vert\vert$, with $\epsilon$ a noise modelling term. 

\begin{figure}
\centering
\includegraphics[width=0.9\linewidth]{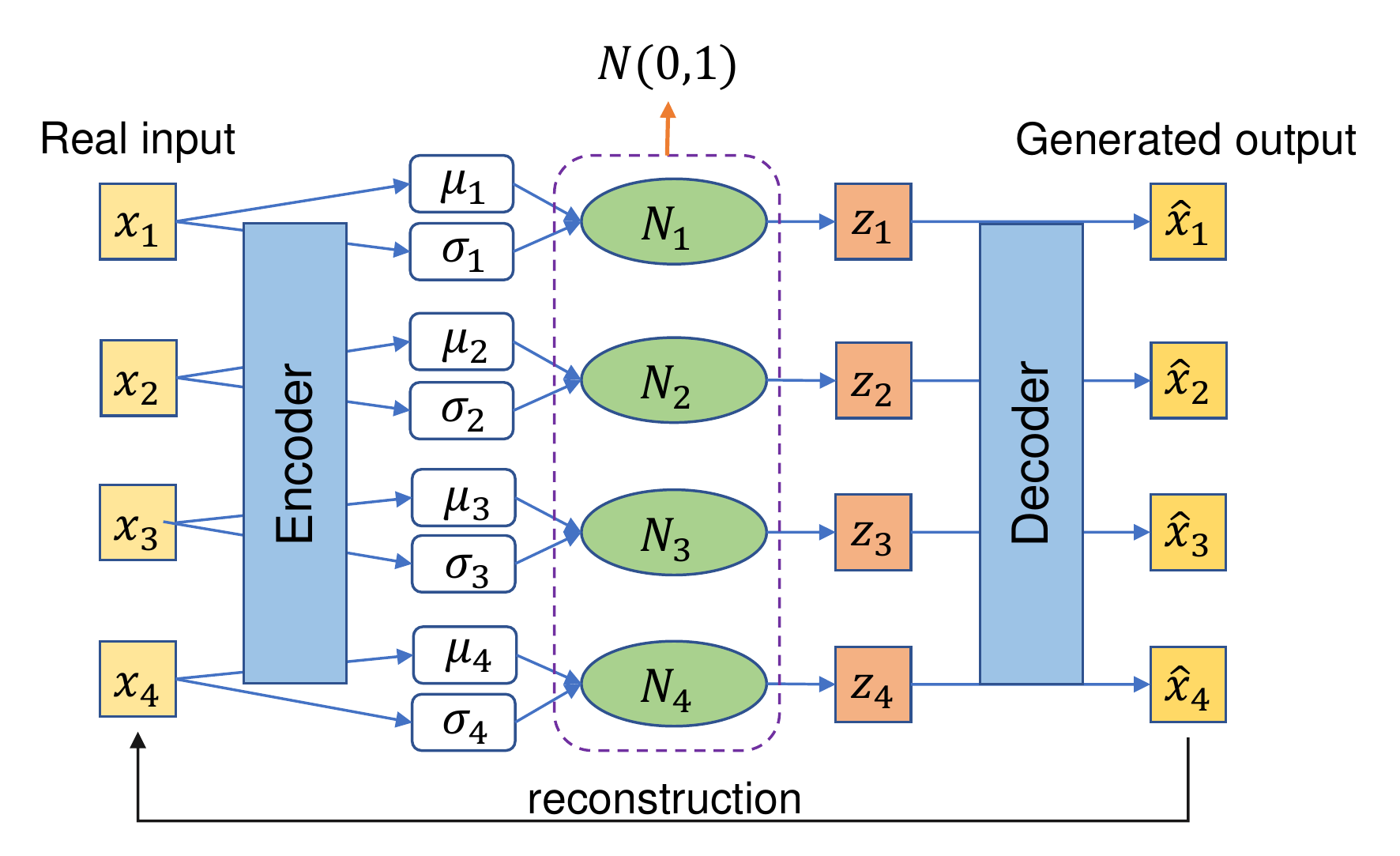}
\caption[vae]{Variational AutoEncoder (VAE)~\cite{kingma2013auto}. Instead of encoding the input $X$ to a fixed latent vector, VAE maps input $(x_1,x_2,\dots)$ to a multi-dimensional Gaussian distribution with non-zero mean $\mu=(\mu_1,\mu_2,\dots)$ and diagonal covariance matrix $\Sigma=\text{diag}(\sigma_1^2,\sigma_2^2,\dots)$. Reconstruction works through decoding sampled latent vectors $(z_1,z_2,\dots)$ from this distribution.}
\label{fig:vae}
\end{figure}

VAE decouples the encoder and the decoder by encoding a latent distribution. For each input sample $x_i$, instead of a single deterministic latent representation $z_i$, the encoder $E$ generates two latent vectors $\mu_i$ and $\sigma_i$ representing the mean and variance of a latent Gaussian distribution. The decoder then randomly samples a representation vector $z_i$ sampled from the Gaussian distribution $\mathcal{N}(\mu_i,\sigma_i)\sim\exp\left[-(z_i-\mu_i)^2/\sigma_i^2\right]$ to reconstruct $x_i$, (see Fig. \ref{fig:vae}).

The required mathematical machinery to properly define the VAE training loss builds on \textit{variational inference}~\cite{blei2017variational}. However, in practice, it amounts to adding a loss term that forces the latent distribution to a unit Gaussian, i.e., $\mathcal{N}(\mu_i,\sigma_i)\to\mathcal{N}(0,1)$ for all training samples $x_i$. This can be realized by minimizing the \textit{KL divergence} between the two distributions:

\begin{equation}
KL(N(\mu, \sigma^{2}) \| N(0,1))=\frac{1}{2}(-\log \sigma^{2}+\mu^{2}+\sigma^{2}-1)
\end{equation}

\noindent Together with the reconstruction term, the total loss  of the VAE reads:

\begin{equation}
\label{eq:VAE}
\mathcal{L}_\text{VAE}=
    \left\vert x-D(E(x))\right\vert + \lambda(-\log \sigma^{2}+\mu^{2}+\sigma^{2}-1)
\end{equation}

\noindent where $\lambda$ is a weighting parameter.

Most recently, masked autoencoder (MAE) \cite{he2021masked} raised great attention in the computer vision community with a breakthrough in autoencoding self-supervised pre-training of vision transformers. Inspired by denoising autoencoder, MAE masks out random patches of the input image, sends visible patches to the encoder, and reconstructs the missing patches from the latent representation and masked tokens. This work proves the potential of transformer-based autoencoders for self-supervised visual representation learning.

\textbf{Remote sensing.} Autoencoders have been widely used to learn representation from various remote sensing data like multispectral images~\cite{lu2017remote,zhang2016change,sharma2021self,madhuanand2021self,peng2020urban,nguyen2021self,tomenotti2020heterogeneous,palsson2022blind}, hyperspectral images~\cite{kemker2017self,mou2017unsupervised,imamura2019self,hong2021endmember,yao2020cross,chen2021hyperspectral,li2021self,ozkan2018endnet} and SAR images~\cite{hughes2018mining,molini2021speckle2void,yuan2019blind,pu2022sae}.
Lu \etal~\cite{lu2017remote} proposed a combination of a shallowly weighted de-convolution network
with a spatial pyramid model in order to learn multi-layer feature maps and filters for input images.
In a subsequent step, these get classified by a support vector machine (SVM) for scene classification.
Zhang \etal~\cite{zhang2016change} utilized a stacked denoising autoencoder to learn image features for
change detection of imagery with various spatial resolutions. 
\textcolor{blue}{In hyperspectral image analysis, autoencoders are either exploited for pre-training~\cite{mou2017unsupervised}, or become ingredients in common downstream tasks, such as hyperspectral image classification~\cite{kemker2017self}, hyperspectral image denoising~\cite{9383811}, hyperspectral image restoration~\cite{imamura2019self, qian2021hyperspectral}, and hyperspectral image unmixing~\cite{hong2021endmember,palsson2022blind,ozkan2018endnet}. }

\subsubsection{Generative Adversarial Networks (GAN)}

Generative adversarial networks~\cite{goodfellow2014generative,radford2015unsupervised} proposed the \textit{adversarial training framework}---a strategy ruled by minimax optimization~\cite{bertsimas2013robust}. GAN training may be viewed as a two players game: the generator $\mathcal{G}$ generates fake samples $\overline{x}=\mathcal{G}(z)$ from random latent vector $z$, and the discriminator $\mathcal{D}$ aims to distinguish $\overline{x}$ from real data samples $x$. 

When the discriminator's output is interpreted as probability distribution, i.e.\ $\mathcal{D}(x)\in[0,1]$, a widely adopted, unified form of the GAN-loss reads
\begin{equation}
\max _{\mathcal{D}} \min _{\mathcal{G}} \mathcal{L}_\text{GAN} = \log \mathcal{D}(x) +\log[1-\mathcal{D}(\mathcal{G}(z))]
\end{equation}
where $x$ represents a real sample and $\mathcal{G}(z)$ represents a fake sample. During training, $\mathcal{G}$ is optimized to fool the discriminator by maximizing $\mathcal{D}(\mathcal{G}(z))$ for any $z$, while $\mathcal{D}$ is tuned to minimize $\mathcal{D}(\overline{x})$ for fake data $\overline{x}$ and maximize $\mathcal{D}(x)$ for real data $x$. Accordingly, the parameters of $\mathcal{G}$ and $\mathcal{D}$ get simultaneously tuned to minimize and maximize $\mathcal{L}_\text{GAN}$, respectively.

Contrarily to autoencoders where $z=E(x)$, a GAN’s latent representation $z$ is implicitly modeled through the inverse of the generator, $\mathcal{G}^{-1}$. To approximate $\mathcal{G}^{-1}$ and obtain $z$ for a given sample $x$, \textit{adversarial autoencoders} (AAE)~\cite{makhzani2015adversarial} can be used. For real samples $x$, $\mathcal{G}$ gets fed by $z=E(x)$ of a (trained) autoencoder $D\circ E$. The discriminator contributes to the loss by distinguishing fakes $\overline{x}=\mathcal{G}(z)$ from real samples $\tilde{x}=\mathcal{G}(E(x))$. Once $\mathcal{G}$, $\mathcal{D}$, and $D\circ E$ are simultaneously optimized, $E$ serves to encode data points $x$ into learnt representations $z$.

\begin{figure}
\centering
\includegraphics[width=0.9\linewidth]{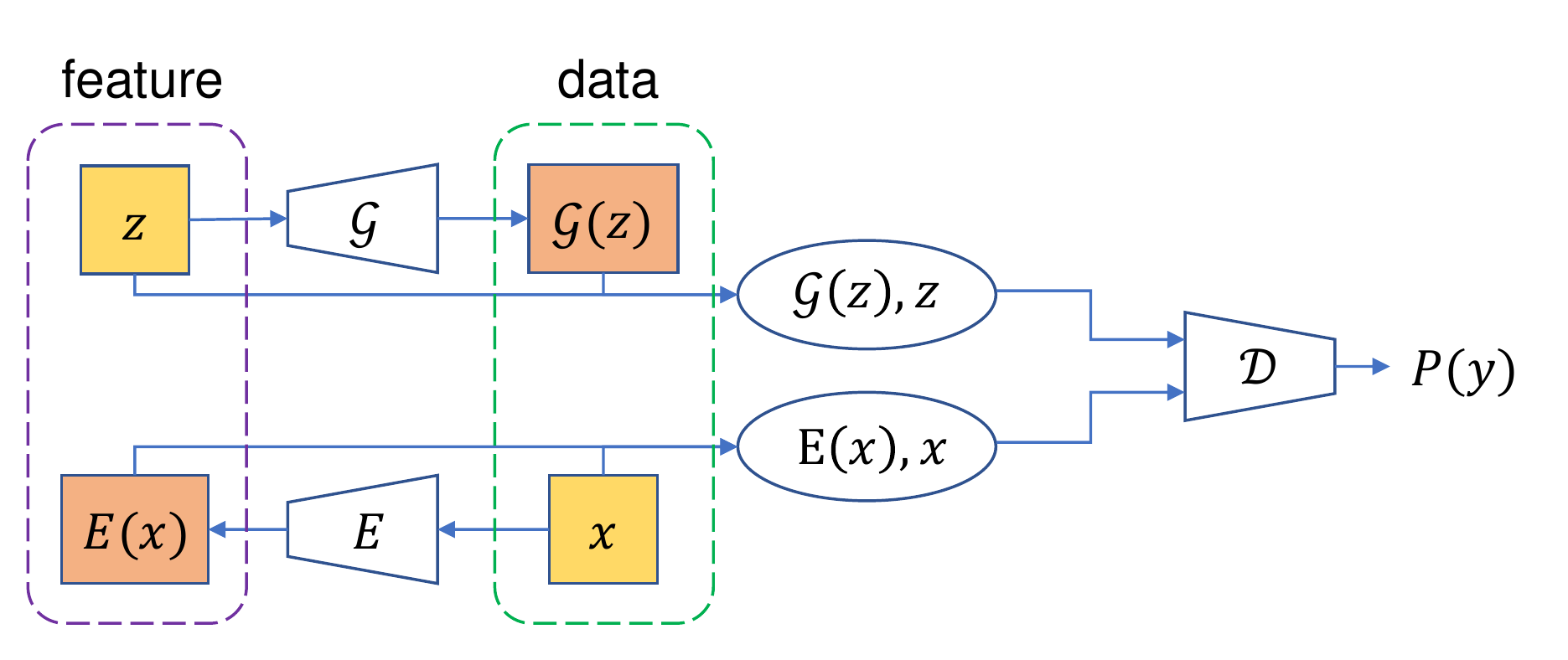}
\caption[BiGAN]{Bidirectional Generative Adversarial Networks (BiGAN)~\cite{donahue2016adversarial}. BiGAN includes an encoder $E$ which maps data $x$ to latent representations $z$. The BiGAN discriminator $\mathcal{D}$ jointly acts in data and latent space: $x$ versus $\mathcal{G}(z)$), and $E(x)$ versus $z$, respectively.}
\label{fig:BiGAN}
\end{figure}

To make GAN's training scheme more symmetric, \textit{BiGAN}~\cite{donahue2016adversarial,donahue2019large} and \textit{ALI}~\cite{dumoulin2016adversarially} have been proposed. As depicted by Fig. \ref{fig:BiGAN}, in addition to generating fake data $\overline{x}=\mathcal{G}(z)$ through $\mathcal{G}$, an encoder $E$ generates latent space representations $\overline{z}=\mathcal{E}(x)$. A discriminator $\mathcal{D}=\mathcal{D}(x,z)$ is trained to distinguish tuples of fakes $(\overline{x},\overline{z})$ from real ones, $(x,z)$.

\textbf{Remote sensing.} While GAN-based methods in remote sensing are few for self-supervised pre-training, several works tend to integrate such methods to target applications.
\textcolor{blue}{Zhu \etal~\cite{zhu2018generative} presented a first work to explore the potential of GAN for hyperspectral image classification. Jin \etal~\cite{jin2021adversarial} proposed to use adversarial autoencoder for unsupervised hyperspectral unmixing.} 
Hughes \etal~\cite{hughes2018mining} proposed a GAN-based framework to generate similar, but novel samples from a given image. Subsequently, these are defined as \textit{hard-negative} samples to match imagery derived from radar and optical sensors. Also, replacing the generator branch of the GAN with a variational auto-encoder improves the quality of negative samples. In another study, Alvarez \etal~\cite{alvarez2020s2} employed the discriminator $\mathcal{D}$ of a GAN for binary change detection, while the generator $\mathcal{G}$ is optimized to induce the distribution of unaltered samples. Walter \etal~\cite{walter2020self} experimented with and evaluate the performance of the BiGAN approach for remote sensing image retrieval based on the similarity of image features. 
\textcolor{blue}{Cheng \etal~\cite{cheng2021perturbation} proposed perturbation-seeking GAN in a defense framework for remote sensing image scene classification. Ozkan \etal~\cite{ozkan2020spectral} proposed to use Wasserstein GAN loss to optimize the multinomial mixture
model for hyperspectral unmixing.}

\subsection{Predictive Methods}

While most generative methods perform pixel-level reconstruction or generation, predictive self-supervised methods are based on auto-generated labels. In fact, one may argue that being able to generate very high dimensional data points, such as images, is not necessary to learn useful representations for many downstream tasks. Instead, one can focus on predicting specific properties of the data, which is the general idea behind predictive methods. To this end, the so-called \textit{pretext tasks}\footnote{We note that the term \textit{pretext task} can also represent any generic objective associated with self-supervised methods. In this paper, we restrict such notion to predictive approaches (where the term originally came from) to make the taxonomy more clear.} get utilized. A predictive method firstly designs a suitable pretext task for the dataset, prepares self-generated labels, and trains a model to predict such labels and learn data representations.

Predictive self-supervised learning targets two possible drawbacks associated with generative methods that perform pixel-level reconstruction: $(1)$ pixel-level loss functions may overly focus on low-level details whereas in practice such details are irrelevant for a human to recognize the contents of an image, $(2)$ pixel-based reconstruction typically do not involve pixel-to-pixel (long-range) correlations that can be important for image understanding. Based on the assumption that providing the network with relevant high-level pretext tasks, the network may learn high-level semantic information. 

In general, the design of different pretext tasks harnesses various context information of the input data. According to distinct context attributes, we categorize pretext tasks as follows: spatial context, spectral context, temporal context, and other semantic contexts.

\begin{figure*}[htp]
\centering
\includegraphics[width=0.8\textwidth]{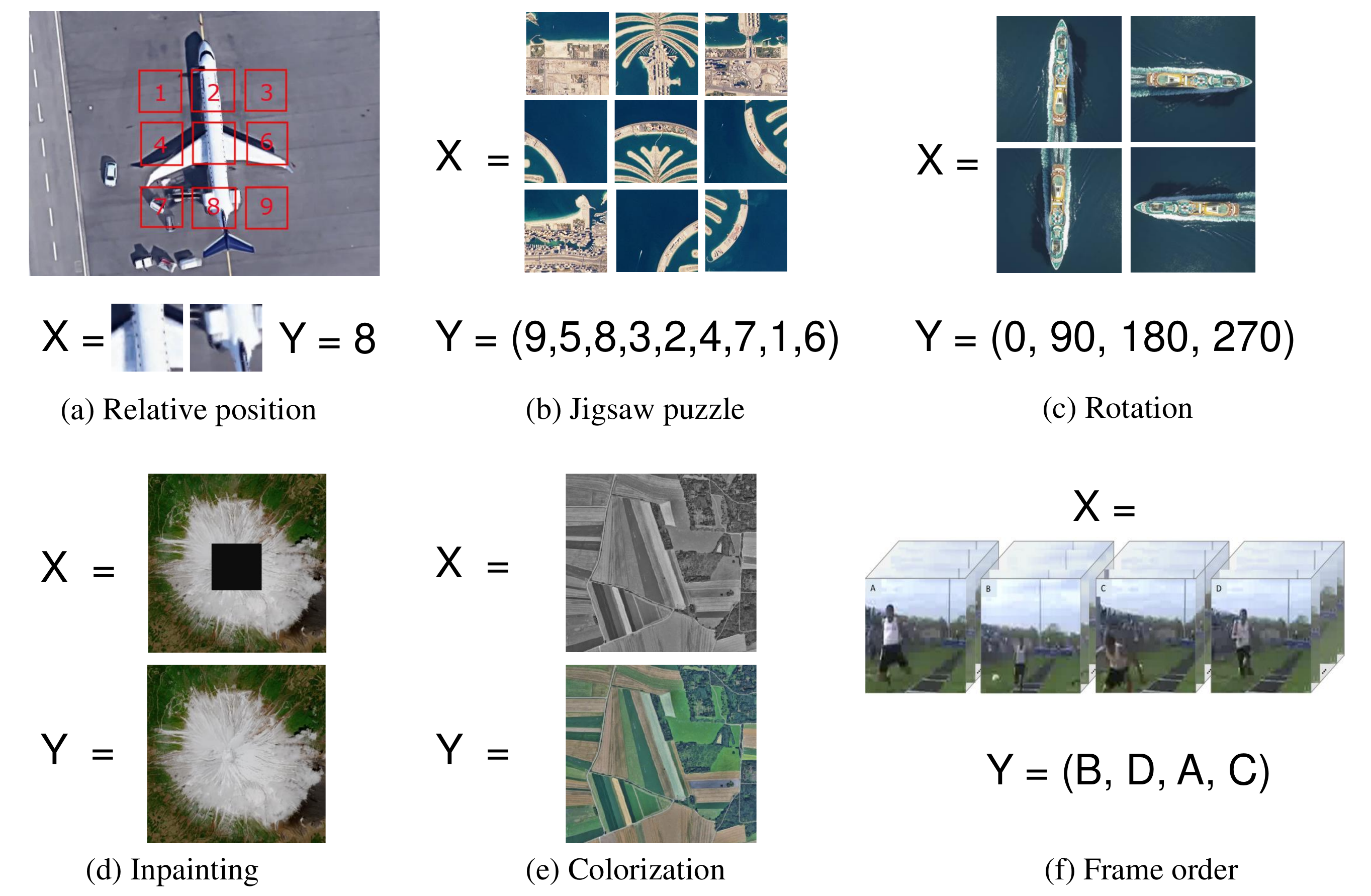}

\caption[]{%
\textcolor{blue}{
    Examples of different predictive pretext tasks. (a),(b),(c), and (d) are based on spatial contexts;
    (e) is based on spectral context; (f) illustrates temporal context.
}

}
\label{fig:pretext-tasks}
\end{figure*}

\subsubsection{Spatial Context}

Images contain rich spatial information for designing self-supervised pretext tasks. Doersch \etal~\cite{doersch2015unsupervised} proposed the first example of such methods, predicting the relative position of pairs of randomly cropped image patches drawn from a given input sample. It is assumed that doing well on this task requires a global understanding of the scene and its objects contained. Accordingly, a valuable visual representation is expected to extract the composition of objects in order to reason about their relative spatial location. 

Following this paradigm, the literature reveals a multitude of methods to learn image features solving an increasingly complex set of spatial puzzles
\cite{noroozi2016unsupervised,wei2019iterative,noroozi2018boosting,santa2018visual,Chen_2021_CVPR}. For example, Noroozi \etal~\cite{noroozi2016unsupervised} built a jigsaw puzzle from randomly ordered tiles of an image. The network was then trained to predict the correct order of tiles. However, given 9 image patches, there are $9!=362,880$ distinct permutations and a network is unlikely to recognize all of them due to visual ambiguity. To limit the number of permutations, the
\textit{Hamming distance}~\cite{6772729} was employed to pick a subset of permutations that are significantly diverse. Borrowed from information theory, here the Hamming distance essentially measures the minimum number of tile permutations required in order to recover the image. Further works building on this approach tried to improve the jigsaw puzzles baseline~\cite{wei2019iterative} by introducing tiles from other images~\cite{noroozi2018boosting,Chen_2021_CVPR}, and by
extending to larger-size puzzles~\cite{santa2018visual}. Along this line, one important thing to be taken into account is that if carelessly designed, the model can "cheat" from low-level details like edges.

Another set of spatial-context-based pretext tasks exploit geometric transformations of input imagery. Among others, predicting rotation angles
\cite{gidaris2018unsupervised,feng2019self} and image inpainting~\cite{pathak2016context,iizuka2017globally}
are popular approaches. In~\cite{gidaris2018unsupervised}, the input image was transformed by four separate rotation angles, which then served as the label for the network to predict.~\cite{pathak2016context} cut a patch from the input for the network to recover. We note that this set of image inpainting strategies have also close relation to generative self-supervised methods---as discussed in Fig. \ref{sec:GenerativeMethods}.

\textbf{Remote sensing.} Zhao \etal~\cite{zhao2020self} proposed a multi-task framework to simultaneously learn from rotation pretext and scene classification to distill task-specific features adopting a semi-supervised perspective. Zhang \etal~\cite{zhang2019rotation} proposed to predict a set of rotation angles from a sequence of rotated SAR-probed targets for object detection. Singh \etal~\cite{singh2018self} utilized image inpainting as a pretext task for semantic segmentation arguing for spatial correlation of the pretext task to the downstream task. In~\cite{tao2020remote} the authors exercised both relative position and inpainting as pretext tasks for remote sensing scene classification. The study added a contrastive self-supervised component referred to as \textit{instance discrimination} which
we discuss in the following section. \textcolor{blue}{Ji \etal~\cite{ji2022few} tackled the few-shot scene classification problem by incorporating two pretext tasks into training: rotation prediction and contrastive prediction. An additional adversarial model perturbation term is also used for regularization.} 

Notably, jigsaw puzzles are rarely leveraged in remote sensing. Potentially, spatial correlation in overhead imagery is less dominant. Indeed, translational invariance is prominent in blocks of urban areas, across water surfaces, and many other kinds of natural scenes (desert, forest, mountain ranges, etc.).

\begin{figure*}
\centering
\includegraphics[width=0.70\linewidth]{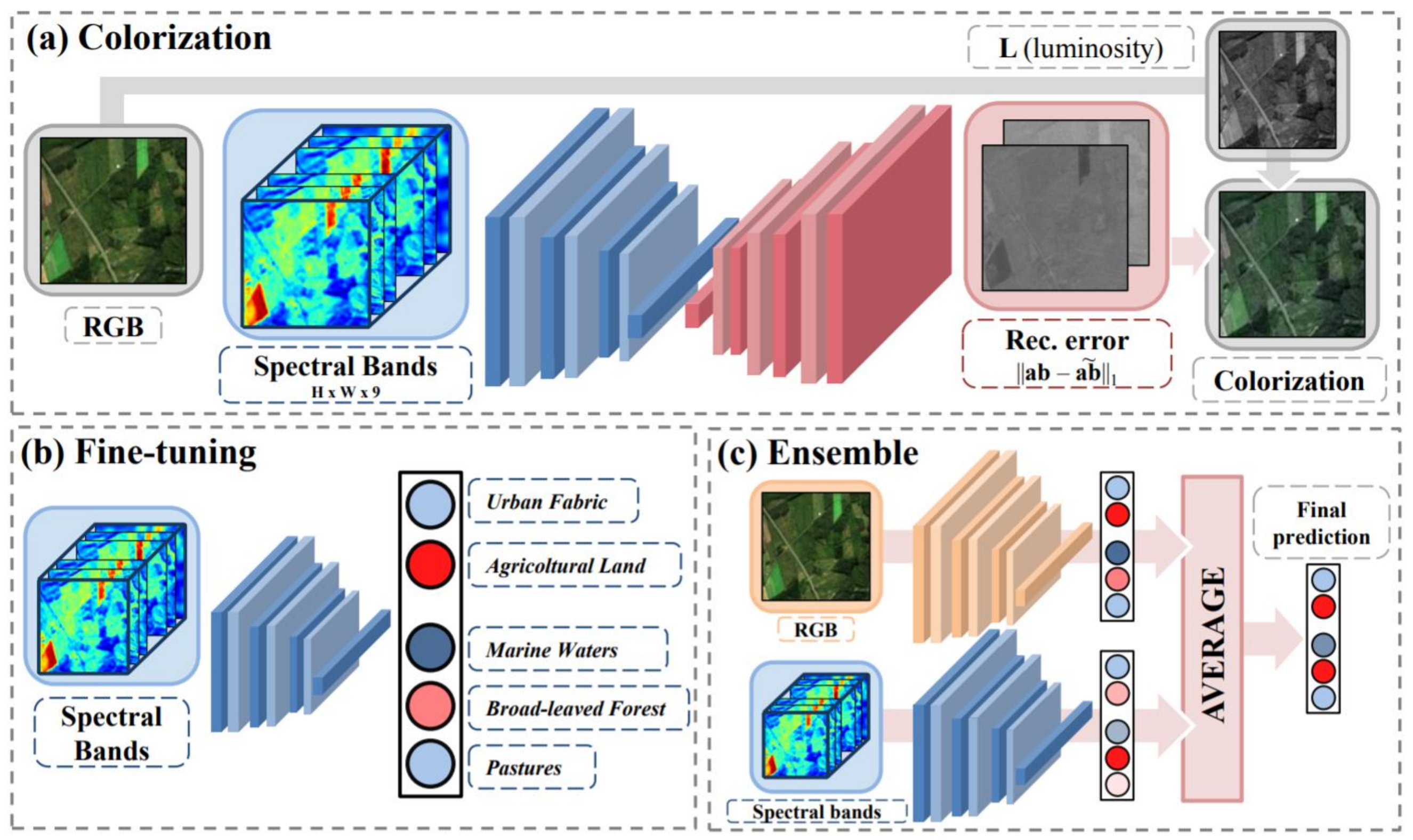}
\caption{The color out of space~\cite{vincenzi2021color}. CIELAB-encoded RGB imagery of a scene is predicted from other spectral bands for the extraction of representations. ©[2021] IEEE.}
\label{fig:color_out_of_space}
\end{figure*}

\subsubsection{Spectral Context}

Spectral information is another basis for the design of pretext tasks which is inspired by image colorization in computer vision. Zhang \etal~\cite{zhang2016colorful} proposed one of the first of such tasks: a self-supervised network learns channel-specific representations by predicting a spectral channel taking other channels as input in \textit{CIELAB} color space~\cite{hill1997comparative}. It turned out that the effectiveness of the color space roots in encoding according to human perception: the distance of two points reflects the amount of visually perceived change of the corresponding colors. The work was further extended to hue/chroma color space, and per-pixel color histograms were applied for better representation learning~\cite{larsson2016learning}. While these two works learn the representation of a single channel (the gray-scale intensity), a cross-channel representation learning method was proposed in~\cite{Zhang_2017_CVPR}. Two sub-encoders got utilized to extract distinctive color channel representations, which are then concatenated to provide full-channel representations. Larsson \etal~\cite{larsson2017colorization} proposed a systematic analysis on image colorization as a pretext task.

\textbf{Remote sensing.} Given remote sensing imagery contains spectral bands beyond the standard RGB color space, there may not exist straightforward extensions to the multi/hyper-spectral domain based on methods established in computer vision. Indeed, sensing data with increased spectral resolution imprints fine details of the physical properties of the earth's surface. Thus, designing pretext tasks related to spectral bands is a subtle exercise to be treated with scientific care. A recent work established by Vincenzi \etal~\cite{vincenzi2021color} provided an initial attempt to leverage spectral context for self-supervised learning (Fig. \ref{fig:color_out_of_space}). Close to an autoencoder, it predicts CIELAB-encoded RGB imagery of a scene from multi-spectral bands for extraction of representations $z_s$. At the same time, a second encoder pre-trained on ImageNet generates representations $z_c$ from the RGB imagery. Downstream classification tasks fine-tune corresponding linear layers $L_{s/c}$ such that the final prediction reads $[L_c(z_c)+L_s(z_s))]/2$. \textcolor{blue}{Wu \etal~\cite{wu2021hyper} perform hyperspectral dimensionality reduction using self-supervised learning by training a model to predict low-dimensional representations generated by classic nonlinear manifold embedding methods (e.g., LLE).}

\subsubsection{Temporal Context}

Among others, temporal context is very important, for example, in video understanding. Without the cut, footage clearly carries temporal correlation from frame to frame. With a similarity to both spatial and spectral predictive methods, temporal pretext tasks can be designed in two ways: $(1)$ shuffling timestamps of frames to let a neural network predict the correct sequence, $(2)$ masking one or several frames for the network to predict the missing frame.

Missing frame prediction in self-supervised learning typically estimates future snapshots from a short clip of video recordings. Provided its superior performance to model temporal dynamics, LSTM~\cite{6772729} or LSTM variants dominate existing methods to encode temporal correlation in video data~\cite{srivastava2015unsupervised,villegas2017decomposing,finn2016unsupervised}.

Frame order prediction in self-supervised learning can be further divided into two distinct approaches: $(1)$ \textit{temporal order verification}~\cite{misra2016shuffle,wei2018learning,fernando2017self,misra2016unsupervised}, and $(2)$ \textit{temporal order recognition}~\cite{lee2017unsupervised,xu2019self,kim2019self}. While the former amounts to binary classification whether or not a sequence of frames is in temporal order, the latter goes beyond by explicitly assigning timestamps to a set of given frames. 

One important thing to note for video understanding is that efficient self-supervised training of footage data may involve various steps of preprocessing. For example, the frames of a video may have different importance for the understanding of the event. Along this line, frame sampling strategies play a big role to boost the performance of temporal pretext tasks. A representative method was proposed by Misra\etal~\cite{misra2016shuffle} that sampled video recordings from frames with significant motion as indicated by the magnitude of the \textit{optical flow}~\cite{fortun2015optical}.
 
\textbf{Remote sensing.} Though having a promising future, satellite-based video recording is not yet common in remote sensing. However, temporal stamps of remote sensing data are very important for applications like change detection or crop type classification. Dong \etal~\cite{dong2020self} quantified temporal context by coherence in time, proposing a self-supervised representation learning technique for remote sensing change detection. The model gets optimized to identify sample patches in two snapshots of the same geospatial area. The identification network (snapshot one vs.\ two) is designed to imitate the discriminator $\mathcal{D}$ of generative adversarial networks with $\mathcal{G}$ generating the self-supervised data representation, as displayed in \textcolor{blue}{Section} \ref{sec:GenerativeMethods}. The method yields improved, robust differentiation for change detection. \textcolor{blue}{Yuan \etal \cite{yuan2020self} proposed a Transformer-based self-supervised methods for satellite time series classification. By predicting randomly contaminated observations given an entire time series of a pixel, the model is trained to leverage the inherent temporal structure of satellite time series to learn general-purpose spectral-temporal representations. The work was further improved in \cite{yuan2022sits}, where the network is asked to regress the central pixels of the masked patches for patch-based representation learning.}

\subsubsection{Other Semantic Contexts}

Apart from the above three types of contexts, there are also other semantic contexts that can be seen using integrated information of the above-mentioned contexts. Noroozi \etal~\cite{noroozi2017representation} proposed an artificial supervision signal based on counting visual primitives from an input image and the sum of primitives from cropped tiles. Jenni \etal~\cite{jenni2018self} proposed a self-supervised learning method based on spotting and predicting artifacts in an adversarial manner. To generate images with artifacts, the authors pre-train a high-capacity autoencoder and then implement a damage-and-repair strategy. A discriminator is finally trained to distinguish artifact images and predict what entries in the feature were dropped when damaging and repairing.

Multi-sensor or multi-modal self-supervised learning gathers another set of semantic contexts. Owens \etal~\cite{owens2016ambient} proposed to predict a statistical summary of the sound associated with a video frame. By defining explicit sound categories, the authors formulate this visual recognition problem as a classification task. Ren \etal~\cite{ren2018cross} trained an encoder-decoder network to predict depth map, surface normal map, and instance contour map from a synthetic image, and a discriminator network is trained to distinguish between features from the synthetic image and the real image.

With various pretext tasks proven to be useful for self-supervised representation learning, there's also a trend of gathering different pretext tasks together. A direct example can be seen in~\cite{doersch2017multi}, where the authors investigated the combination of four self-supervised tasks: relative position~\cite{doersch2015unsupervised}, colorization~\cite{zhang2016colorful}, exemplar~\cite{dosovitskiy2014discriminative} (creating a pseudo-class from the augmented view of a single image and predicting the class, this is rather a contrastive method which will be discussed in the next section) and motion segmentation~\cite{pathak2017learning} (the network learns to classify which pixels of a single frame will move in subsequent frames). It was shown that combining self-supervised tasks will in general improve the performance and lead to faster training.

\textbf{Remote sensing.} Due to the difference between remote sensing data and common images studied in the computer vision community, there's also a large potential for designing remote sensing-specific pretext tasks. Hermann \etal~\cite{hermann2020self} proposed to predict different views (pose, projection, and depth) as well as reconstruction of the input for monocular depth estimation. 
\textcolor{blue}{Gao \etal~\cite{gao2022self} proposed to train the model to realize envelope data extrapolation from a frequency band to its adjacent low frequency band. He \etal~\cite{huang2022self} proposed to reconstruct the corrupted Seismic data with consecutively missing traces, in which the pseudolabels are automatically created from the uncorrupted parts of the observed data.} 
Ayush \etal~\cite{ayush2020geography} designed a pretext task based on predicting the geo-location of the input image. Li \etal~\cite{li2021geographical} made use of GlobeLand30~\cite{jun2014open}, a global land cover product that divides the earth into different areas and includes ten different land cover types. By aligning the geo-location of the input image with the land cover map from GlobeLand30, the land cover class of the input image is then used as a pretext task for self-supervised learning. However, it has also to be noted that pretexts like geo-location itself are usually too simple to learn a good representation, thus they are often used as auxiliary tasks for self-supervised learning (as in the two examples above). In general, how to design or integrate a suitable pretext task is still an important research question to be explored.

\subsection{Contrastive Methods}
\label{sec:ContrastiveMethods}

\begin{figure}
\centering
\includegraphics[width=0.75\linewidth]{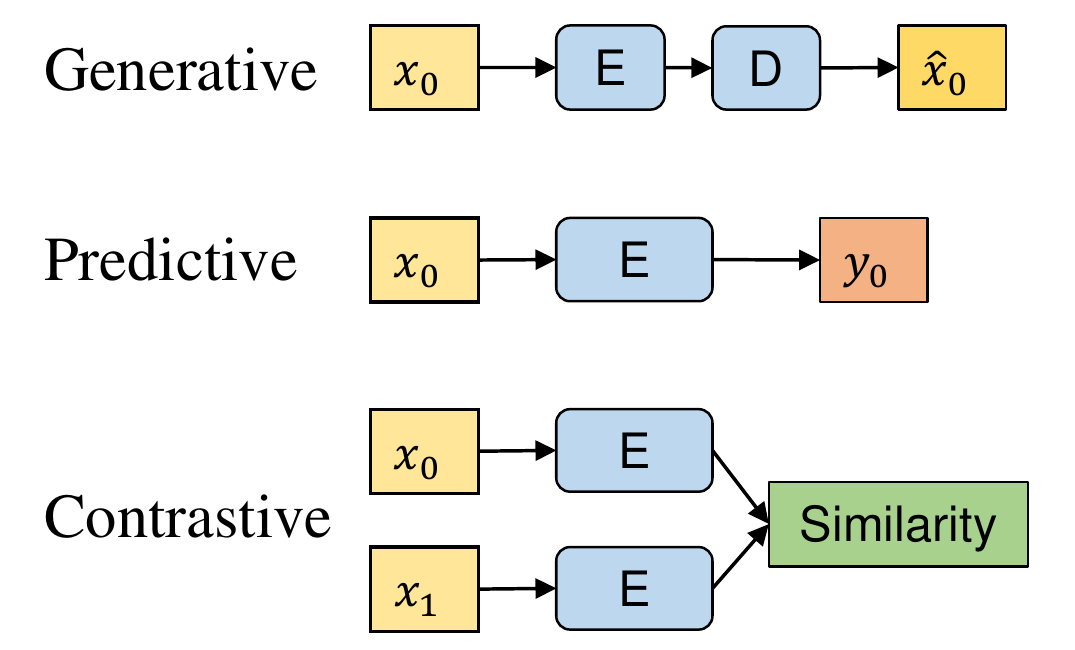}
\caption[GPC]{A comparison of the general structures between generative, predictive, and contrastive self-supervised learning. $E$ and $D$ refer to the encoder and decoder network. While generative and predictive methods calculate the loss in the output space, contrastive methods calculate the loss in the representation space.}
\label{fig:GPC}
\end{figure}

The performance of predictive self-supervised learning depends largely on a good pretext task, which is often very difficult to design and may even lead to pretext-specific representation, decreasing the network's generalizability. To tackle this problem, contrastive methods come into play, giving the network more freedom to learn high-level representations that do not rely on a single pretext task. As the name implies, contrastive methods\footnote{In some literature, the term contrastive is only used to denote methods that include negative samples. Yet, we follow the more common terminology in which contrastive methods also include the recent works without negative samples.} train a model by contrasting semantically identical inputs (e.g., two augmented views of the same image) and pushing them to be close-by in the representation space. Therefore, by design, contrastive methods usually follow a common Siamese-like architecture design. A side-by-side comparison with generative and predictive methods is provided in Fig. \ref{fig:GPC}.

However, only enforcing similarity between pairs of input can easily lead to a trivial solution. Indeed, a constant mapping would be a valid solution to the problem since every pair of input would have identical representations, and thus a maximum similarity. This phenomenon is often referred to as \textit{model collapse} and many solutions have been proposed in the SSL literature to mitigate it. In fact, depending on how collapsing is handled, one can form a sub-taxonomy of contrastive self-supervised learning: negative sampling, clustering, knowledge distillation, and redundancy reduction.

\subsubsection{Negative Sampling}

A basic strategy to avoid model collapse is to include and utilize dissimilar samples to have both positive and negative pairs. This strategy is the first one that has been used in contrastive representation learning. For any data point $x$ (the anchor), the encoder $f$ is trained such that:

\begin{equation}
\label{eq: contrastive}
\operatorname{sim}\left(f(x), f\left(x^{+}\right)\right)\gg\operatorname{sim}\left(f(x), f\left(x^{-}\right)\right)
\end{equation}

\noindent where $x^{+}$ is a data point similar to $x$ (positive sample), $x^{-}$ is a data point dissimilar to $x$ (negative sample), and $sim$ represents a metric that measures the similarity between two pairs of features encoded by $f$. Positive samples need to be generated in a way that preserves the semantics of the anchor $x$ (e.g., using data augmentation). Negative samples on the other hand come from other data points in the dataset. The intuition is that, in order to output similar representations for visually different, yet semantically similar inputs, while repulsing negative samples in the embedding space, the network has to learn useful high-level representations of the input data. Once pre-trained, the encoder $f$ can be further transferred to extract representative features of downstream datasets.  

A lot of methods have been developed based on the general objective of Eq. \ref{eq: contrastive}, of which the earliest works~\cite{chopra2005learning,weinberger2009distance,schroff2015facenet} proposed triplet losses using a max-margin approach to separate positive from negative examples:

\begin{equation}
\mathcal{L}\left(x, x^{+}, x^{-}\right)=\max \left(0,\left\|x-x^{+}\right\|_{2}-\left\|x-x^{-}\right\|_{2}+1\right)
\end{equation}

\noindent where $x$, $x^{+}$, $x^{-}$ represent the anchor, the positive sample and the negative sample, respectively. Noroozi \etal~\cite{noroozi2017representation} extended this triplet scheme for better transfer learning performance by solving an additional pretext task: counting the visual primitives of tiled patches.

\begin{figure}
\centering
\includegraphics[width=0.95\linewidth]{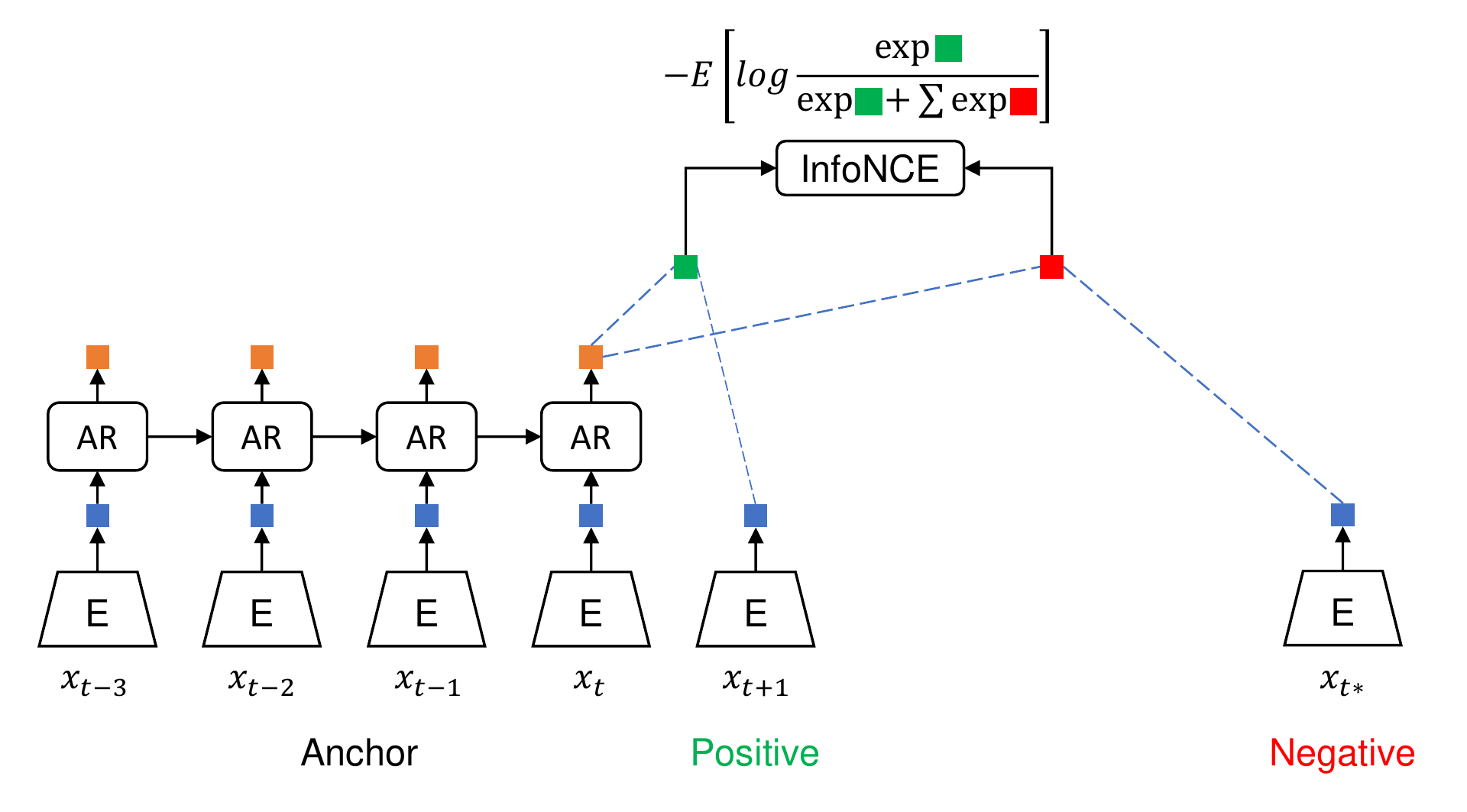}
\caption[CPC]{
\textcolor{blue}{Contrastive Predictive Coding (CPC)~\cite{oord2018representation}. Take time series as an example, the consecutive time stamp is a positive sample, while clips sampled from other scenes are negative samples.}
}
\label{fig:CPC}
\end{figure}

Inspired by noise-contrastive estimation (NCE)~\cite{gutmann2010noise} and word2vec~\cite{mikolov2013efficient,mikolov2013distributed}, Oord \etal~\cite{oord2018representation} started another set of contrastive methods with negative sampling by introducing contrastive predictive coding (CPC) with \textit{InfoNCE} loss:

\begin{equation}
\mathcal{L}_{N}=-\mathbb{E}_{X}\left[\log \frac{\exp \left(f(x)^{T} f\left(x^{+}\right)\right)}{\sum_{j=1}^{N} \exp \left(f(x)^{T} f\left(x_{j}\right)\right)}\right]
\end{equation}

\noindent where $N$ is the number of samples, and the denominator consists of one positive and $N - 1$ negative samples (see Fig. \ref{fig:CPC}). This is indeed the cross-entropy loss for an N-way softmax classifier that classifies positive and negative samples, with the dot product as the score function in Eq. \ref{eq: contrastive}. Take a piece of time series as an example, a consecutive time stamp is a positive sample, while clips randomly sampled from other scenes are negative samples (see Fig. \ref{fig:CPC}). This work was further improved specifically for image recognition in~\cite{henaff2020data}. 

InfoNCE loss is also connected to mutual information, as minimizing the InfoNCE loss can be seen as maximizing a lower bound of the mutual information between $f(x)$ and $f(x^{+})$~\cite{poole2019variational}. This bridges the connection to mutual-information-based contrastive learning which comes out in parallel with CPC. Deep information maximization (DIM)~\cite{hjelm2018learning} for example, learns image representations by classifying whether a pair of global features and local features are from the same image. The global features are the final output of a convolutional encoder and local features are the output of an intermediate layer in the encoder. This work was further improved by enhancing the association of positive pairs in augmented multi-scale DIM (AMDIM)~\cite{bachman2019learning}, where a positive sample is sampled from an augmented view of the input image.

\begin{figure*}
\centering
\includegraphics[width=\linewidth]{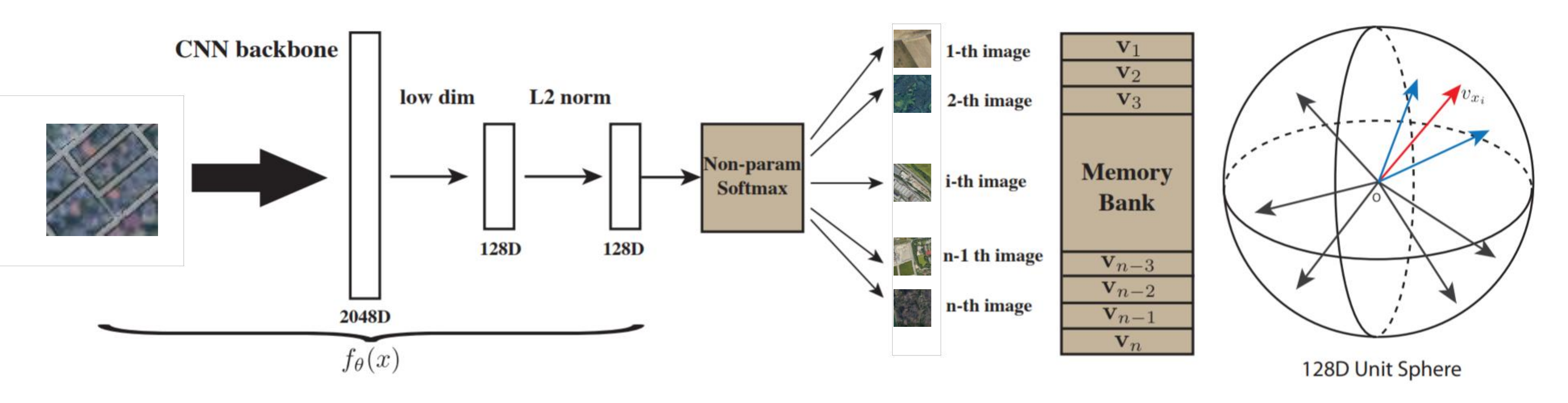}
\caption[InstDisc]{Instance Discrimination (InstDisc)~\cite{wu2018unsupervised}. A backbone CNN is used to encode each image as a feature vector, which is projected to a 128-dimensional space and L2 normalized. The optimal feature embedding is learned via instance-level
discrimination, which tries to maximally scatter the features of training samples over the 128-dimensional unit sphere. ©[2018] IEEE.}
\label{fig:InstDisc}
\end{figure*}

Some further empirical evidence~\cite{tschannen2019mutual} claimed that the success of the models mentioned above (maximizing mutual information) is only loosely connected to mutual information. Instead, an upper bound on mutual information estimator leads to ill-conditioned and lower performance representations. Therefore, more attention should be attributed to the encoder architecture and a negative sampling strategy related to \textit{metric learning}~\cite{kulis2013metric}. Along this line, instance-instance contrastive learning discards mutual information and directly studies the relationships between different samples’ instance-level representations. A prototype work is instance discrimination (InstDisc, see Fig. \ref{fig:InstDisc})~\cite{wu2018unsupervised}, which extended the idea of exemplar~\cite{dosovitskiy2014discriminative} with non-parametric NCE loss (both utilize instance-level discrimination). In fact, both mutual information and instance discrimination are useful for representation learning, and further works tend to combine both of them implicitly.

There are mainly two questions to consider for a contrastive method with negative sampling: (1) what to compare; (2) how to choose positive/negative pairs. Though different methods vary slightly in the contrasting level (mainly context-level like CPC~\cite{oord2018representation} and DIM~\cite{hjelm2018learning} or instance level like InstDisc~\cite{wu2018unsupervised}), what to compare is commonly the encoded features $f(x)$ of positive/negative samples. The choice of positive/negative samples then leads to a range of different methods. While CPC~\cite{oord2018representation} and DIM~\cite{hjelm2018learning} defined features from the same input image as positive pairs and those from other images as negative pairs, AMDIM~\cite{bachman2019learning} extended the choice of the positive sample by sampling from an augmented view of the input image. Contrastive multiview coding (CMC)~\cite{tian2020contrastive} extended the idea to several different views (depth, luminance, luminance, chrominance, surface normal, and semantic labels) of one image and samples another irrelevant image as the negative. This work was further improved by InfoMin~\cite{tian2020makes} studying how to choose best views. In addition, Misra \etal~\cite{misra2020self} proposed pretext-invariant representation learning (PIRL), using pretext tasks in predictive self-supervised learning to transform the input image and let the model learn the invariances.

Contrastive methods tend to work better with more negative examples, since a presumably larger number of negative examples may cover the underlying distribution more effectively. However, the number of negative samples is usually restricted to the size of the mini-batch as the gradients flow back through the encoders of both the positive and negative samples. Consequently, larger mini-batches are preferable, but this in turn is going to be limited by hardware memory constraints. A possible solution is to maintain a separate dictionary, called \textit{memory bank}, which stores and updates the embeddings of samples with the most recent ones at regular intervals. The memory bank contains a feature representation $m_{x}$ for each sample $x$ in dataset $X$. The representation $m_{x}$ is an exponentially moving average of feature representations that were computed in prior epochs. It enables replacing negative samples $m_{x^{-}}$ by their memory bank representations without increasing the batch size. PIRL~\cite{misra2020self} and InstDisc~\cite{wu2018unsupervised} are two representative methods that use a memory bank to boost performance. 

\begin{figure}
\centering
\includegraphics[width=\linewidth]{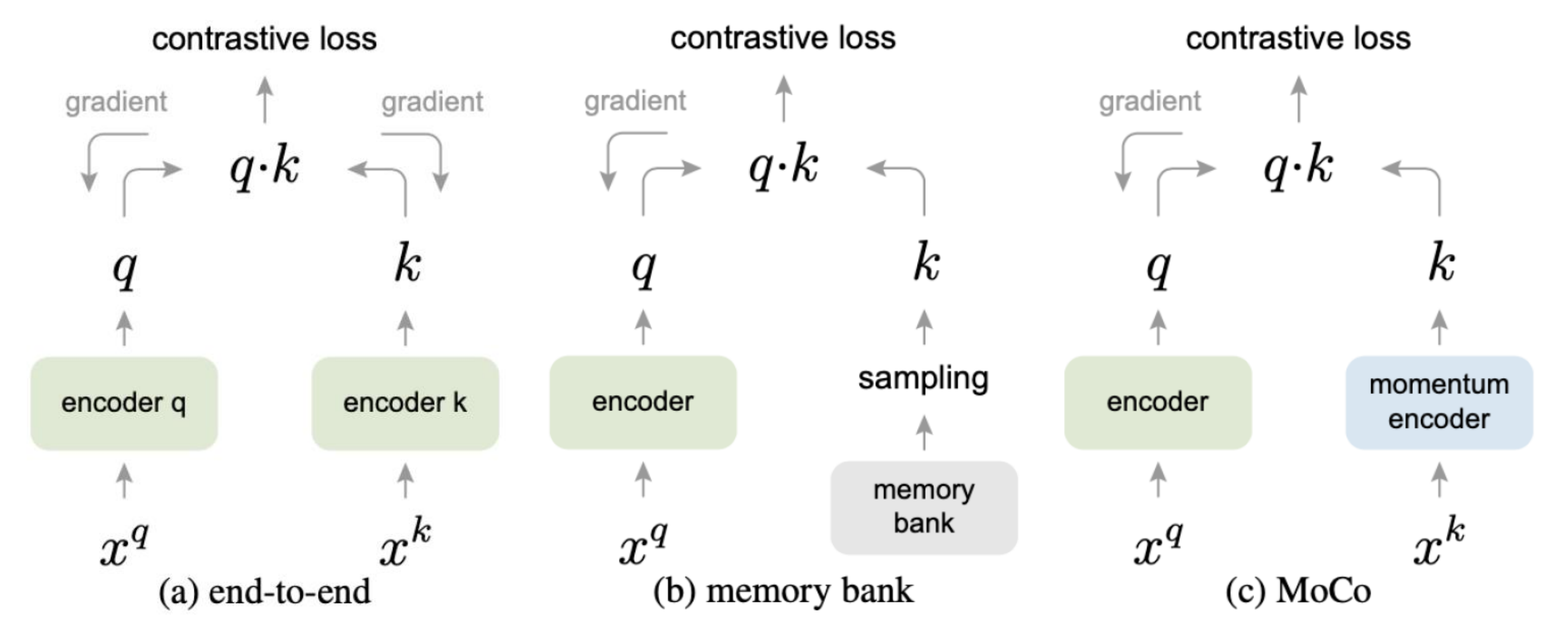}
\caption[moco]{Conceptual comparison of MoCo~\cite{he2020momentum} with two previous contrastive mechanisms. Rather than end-to-end back-propagation or sampling from a memory bank, MoCo encodes the new keys (key representation) on the fly by a momentum-updated encoder and maintains a queue of keys (a dictionary) for storing negative samples. The size of the dictionary can be much larger than a typical mini-batch size, and the samples in the dictionary are progressively replaced. ©[2020] IEEE.}
\label{fig:moco}
\end{figure}

However, maintaining a memory bank during training can be a complicated task, as (1) it can be computationally expensive to update the representations in the memory bank as the representations get outdated quickly in a few passes; (2) the representations in the memory bank are always one step behind the current encoding, which might bring unnecessary mismatches. To address these issues, the memory bank got replaced by a separate module called \textit{momentum encoder} in MoCo~\cite{he2020momentum}. The momentum encoder generates a dictionary as a queue of encoded keys with the current mini-batch enqueued and the oldest mini-batch dequeued. The dictionary keys are defined on the fly by a set of data samples in the batch during training. MoCo also abandons the traditional end-to-end training framework by updating the momentum encoder based on the query encoder (Fig. \ref{fig:moco}):

\begin{equation}
\theta_{k} \leftarrow m \theta_{k}+(1-m) \theta_{q}
\end{equation}

\noindent where $m$ is the momentum coefficient and $k$, $q$ represent key encoder (momentum encoder, no back-propagation) and query encoder (back-propagation) respectively. As a result, it does not require training two separate models and there is no need to maintain a memory bank.

\begin{figure}
\centering

\includegraphics[width=\linewidth]{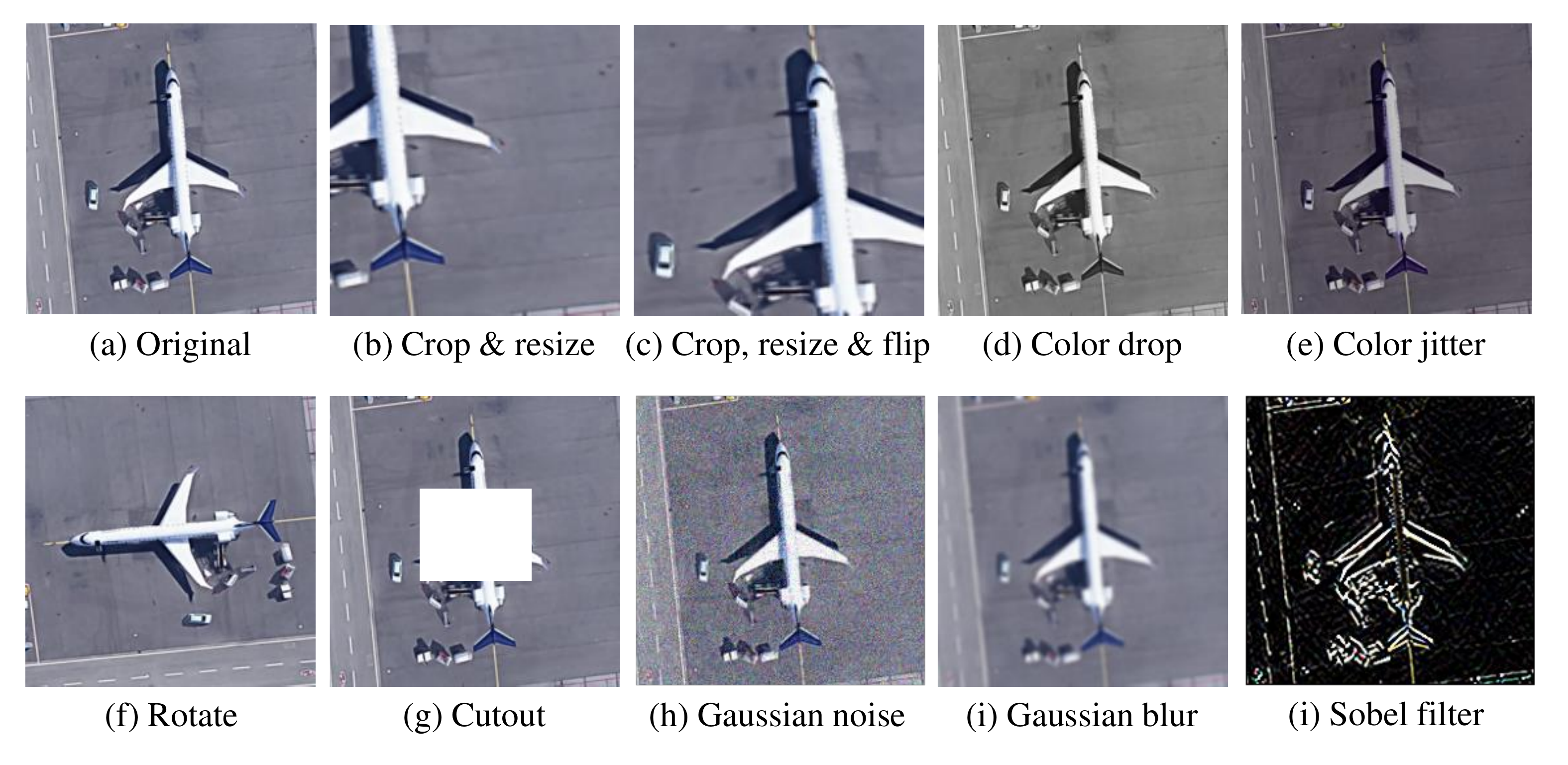}
\caption[dataaug]{
\textcolor{blue}{
Data augmentation operators studied in SimCLR~\cite{chen2020simple}. The selected best group of data augmentations ("random cropping, color jittering, grayscaling, Gaussian blurring and horizontal flipping") are widely referenced in following self-supervised studies.}
}
\label{fig:dataaug_simclr}
\end{figure}

Based on previous works, Chen \etal~\cite{chen2020simple} proposed a milestone of contrastive learning: A simple framework for contrastive learning of visual representations (SimCLR). SimCLR follows the end-to-end training framework and chooses a batch size as large as 8196 to handle the performance bottleneck of the number of negative samples. The main contribution of SimCLR is that it illustrates the importance of a hard positive sampling strategy by introducing data augmentation in 10 forms, and provides a standard scheme of data augmentation for most further works (see Fig. \ref{fig:dataaug_simclr}). SimCLR also provides some other practical techniques for contrastive learning, including an additional learnable nonlinear transformation between the representation and the contrastive loss (the projection head), doubling negative samples and more training steps. Following the strategies of SimCLR, the authors of MoCo proposed an improved version MoCo-v2~\cite{chen2020improved} by improving the data augmentation, adding a projection head, and adapting cosine decay on the learning rate. The authors also proposed a transformer version MoCo-v3~\cite{chen2021empirical} by replacing the ResNet~\cite{he2016deep} backbone of the encoder to a vision transformer~\cite{dosovitskiy2020image}. Meanwhile, the authors of SimCLR improved their model to a second version SimCLR-v2~\cite{chen2020big} for better performance on semi-supervised learning using distillation, with larger encoder networks, larger batch size, and deeper projection heads.

\begin{figure}
\centering
\includegraphics[width=0.9\linewidth]{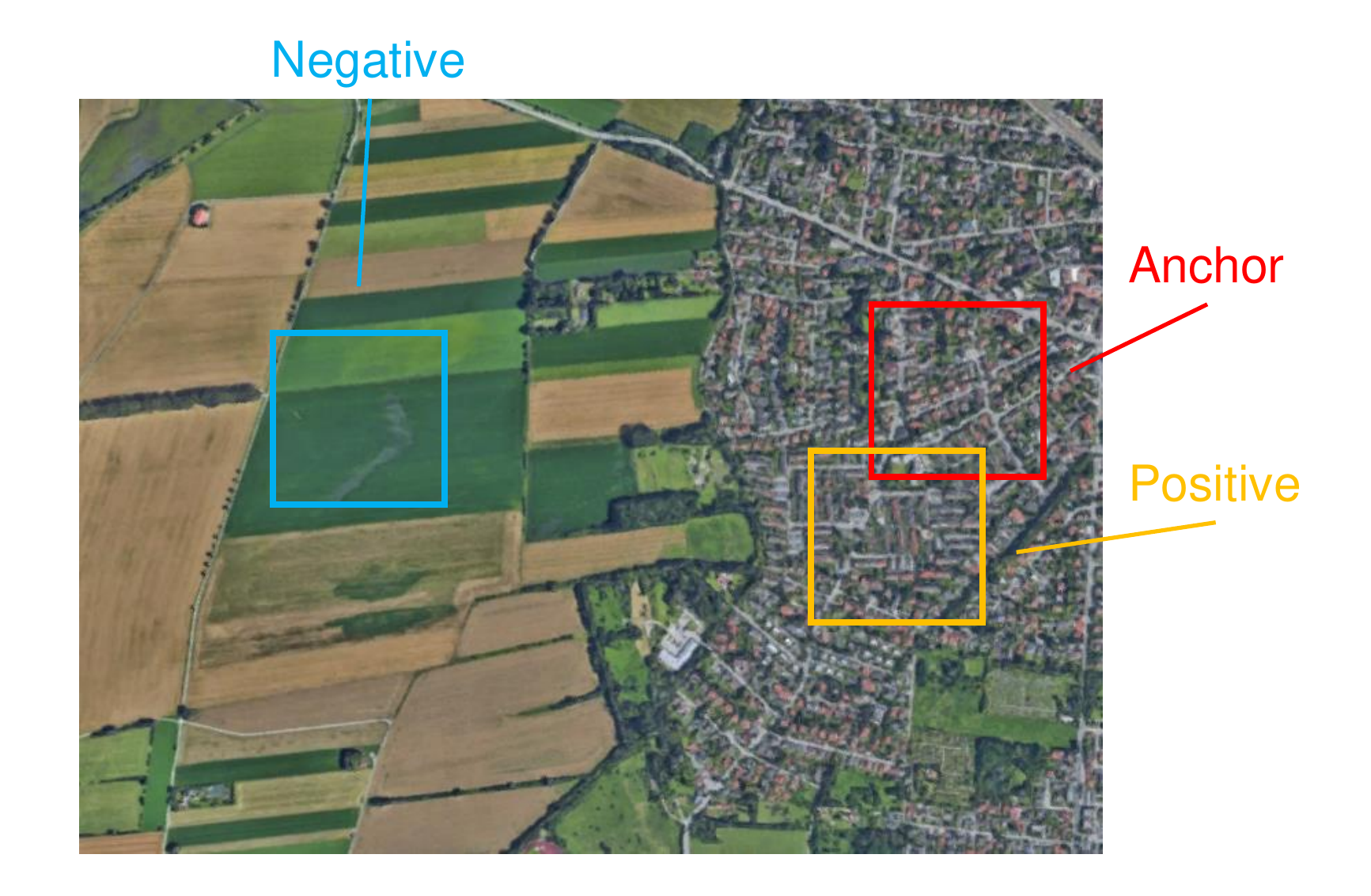}
\caption[tile2vec]{Tile2Vec~\cite{jean2019tile2vec}. A triplet loss is optimized to move neighbour patches close and distant patches far away in the feature space.}
\label{fig:tile2vec}
\end{figure}

\textbf{Remote sensing.} \textcolor{blue}{While contrastive self-supervised learning is relatively new, the wide use of contrastive loss in remote sensing \cite{zhang2022semisupervised,yang2022laboratory,rao2022transferable,jing2022radar,zhang2020unsupervised,geng2022multi} can date back to \cite{cheng2018deep}, where the authors imposed a supervised contrastive regularization term on the CNN features for remote sensing scene classification.} 
The first self-supervised work making use of contrastive learning for remote sensing image representation learning is Tile2Vec proposed by Jean \etal~\cite{jean2019tile2vec}. Similar to CPC~\cite{oord2018representation}, this work was also inspired by word2vec~\cite{mikolov2013efficient,mikolov2013distributed}. The authors trained a convolutional neural network on triplets of tiles, where each triplet consists of an anchor tile $t_{a}$, a neighbor tile $t_{n}$ that is close geographically, and a distant tile $t_{d}$ that is farther away. A triplet loss is used to move closer neighbour tiles and move further distant tiles in feature space. Jung \etal~\cite{jung2021self} later reformulated the triplet loss to binary classification loss and added no-updated fully connected layers to improve robustness. Leenstra \etal~\cite{leenstra2021self} proposed a combination of triplet loss and binary cross-entropy loss (positive pair as 1 and negative pair as 0) for self-supervised change detection.
\textcolor{blue}{Inspired by DIM, Li \etal~\cite{li2022deep} proposed a deep mutual information subspace clustering (DMISC) network for hyperspectral subspace clustering. Hou \etal~ \cite{hou2021hyperspectral} rely on contrastive learning, following a design similar to SimCLR, to pre-train an HSI classification model on a single scene and reduce the need for dense annotations. Contrastive learning is used on the unlabeled pixels with spectral-spatial feature extraction followed by a fine-tuning stage. Similarly, Zhao \etal~ \cite{zhao2022hyperspectral} also apply SimCLR for HSI classification showing promising results with very limited labels. In the same line of works, Zhu \etal~ \cite{zhu2021sc} leverage SimCLR for HSI classification using a multi-scale feature extraction approach. PuHong \etal~ \cite{duan2022self} utilized MoCo for oil spill detection in hyperspectral images. It is worth mentioning that all these hyperspectral data related works rely on simple spatial and spectral augmentations (e.g., random cropping, Gaussian noise, etc.) for view generation. We believe there is room for improvement in this direction.}

Jung \etal~\cite{jung2021contrastive} combined the sampling idea of tile2vec and contrastive architecture of SimCLR, utilizing smoothed representation of three neighbor tiles as a positive sample. Montanaro \etal~\cite{montanaro2021self} proposed to use SimCLR for representation learning of the encoder and perturbation invariant autoencoder for segmentation training of the decoder to perform land cover classification. 
\textcolor{blue}{Scheibenreif \etal~\cite{scheibenreif2022self} tackled land cover classification and segmentation problems using SimCLR with Swin Transformers and by contrasting optical Sentinel 2 and SAR Sentinel 1 patches. Stevenson \etal~\cite{stevenson2022deep} proposed to use SimCLR for representation learning of LiDAR elevation data.} 
Kang \etal~\cite{kang2020deep} define spatial augmentation criteria to uncover semantic relationships among land cover tiles and use MoCo for contrastive self-supervised learning. Li \etal~\cite{li2021remote} added one local matching contrastive loss (between patches within an image) to commonly used global contrastive loss (between different images) to learn rich pixel-level information for semantic segmentation. All above-mentioned methods consider the spatial contexts of remote sensing images and based on that build positive/negative pairs. Apart from these methods, contrastive multi-view coding (CMC) is used in~\cite{stojnic2021self,chen2021self1} and~\cite{liu2020deep} for multispectral and hyperspectral representation learning, and seasonal contrast (SeCo) is proposed in~\cite{manas2021seasonal} to utilize temporal (season) information to create different views. Heidler \etal~\cite{heidler2021self} proposed a combination of triplet loss and all possible pairings (called batch triplet loss) for audio-visual multi-modal self-supervised learning, and Ayush \etal~\cite{ayush2020geography} combined contrastive learning and predictive learning, proposing a combination of MoCo-v2~\cite{chen2020improved} and geo-location as a pretext task for better representation learning of geo-related images. 
\textcolor{blue}{Guan \etal~\cite{guan2022cross} proposed to integrate contrastive NCE loss and masked image modeling for cross-domain hyperspectral image classification.}
   
\subsubsection{Clustering}

Referring back to one of the earliest unsupervised learning algorithms, the series of clustering-based self-supervised methods learn data representation by using a clustering algorithm to group similar features together in the embedding space. In supervised learning, this pulling-near process is accomplished via label supervision; in self-supervised learning, however, we do not have such labels. To solve the problem, DeepCluster~\cite{caron2018deep} proposed to iteratively leverage $K$-means clustering to yield pseudo labels and ask a discriminator to predict the labels. This is also the first work towards clustering-based self-supervised learning.

\begin{figure}
\centering
\includegraphics[width=0.9\linewidth]{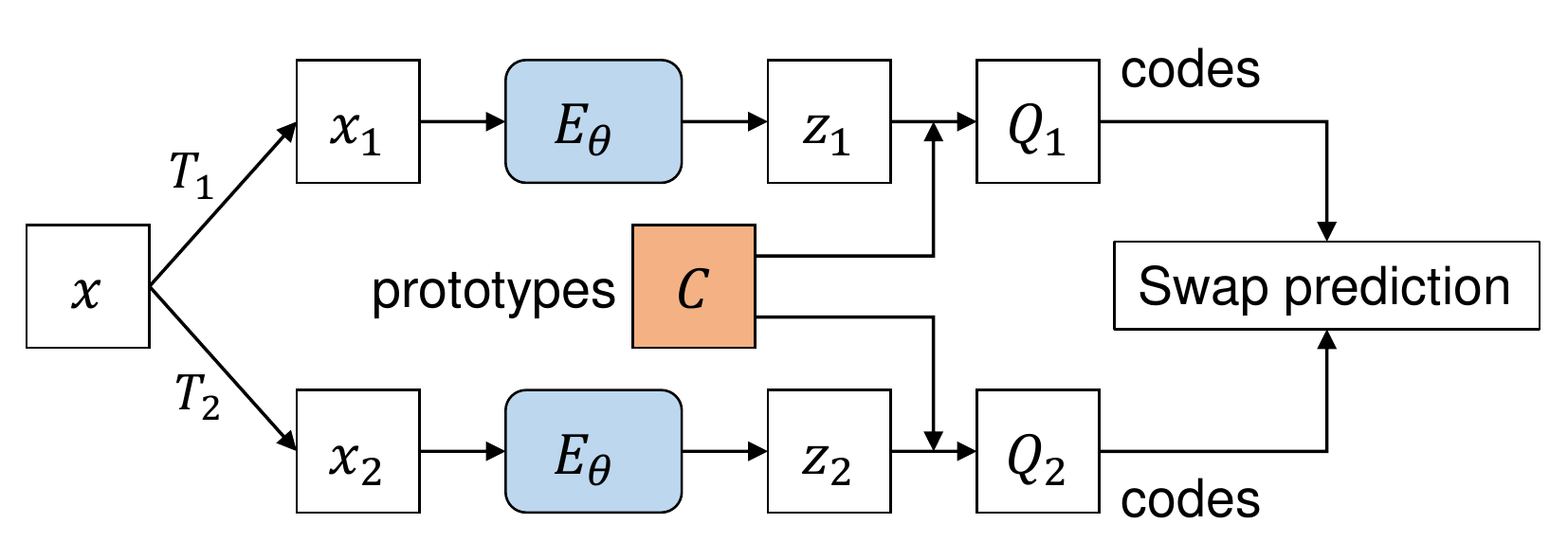}
\caption[SwAV]{Swapping Assignments between Views (SwAV)~\cite{caron2020unsupervised}. “Codes” are obtained by assigning features to learnable prototype vectors, a “swapped” prediction problem is then solved wherein the codes obtained from one data augmented view are predicted using the other view.}
\label{fig:SwAV}
\end{figure}

Self-labeling~\cite{asano2019self} and local aggregation~\cite{zhuang2019local} further pushed forward the boundary of clustering-based methods. Like DeepCluster~\cite{caron2018deep}, local aggregation and self-labeling also use an iterative training procedure, but the specific process within each iteration differs significantly. In self-labeling, instead of $K$-means clustering, an optimal transport problem is solved to obtain the pseudo-labels. In local aggregation, unlike the clustering step of DeepCluster where all examples are divided into mutually-exclusive clusters, the method identifies neighbors separately for each example, allowing for more flexible statistical structures. In addition, local aggregation employs an objective function that directly optimizes a local soft-clustering metric, requiring no extra readout layer and only a small amount of additional computation on top of the feature representation training.

Despite the early success of clustering-based self-supervised learning, the two-stage training paradigm is time-consuming and poor-performing compared to later instance-discrimination-based methods such as SimCLR~\cite{chen2020simple} and MoCo~\cite{he2020momentum}, which have gotten rid of the slow clustering stage and introduced efficient data augmentation strategies to boost the performance. In light of these problems, the authors of DeepCluster~\cite{caron2018deep} brought the idea of online clustering and multi-view data augmentation into the clustering-based contrastive self-supervised approach, proposing to learn features by swapping assignments between multiple views of the same image (SwAV)~\cite{caron2020unsupervised}. The intuition is that, given some (clustered) prototypes, different views of the same images should be assigned to the same prototypes. Meanwhile, an online computing strategy was designed to accelerate code (cluster assignment) computing (see Fig. \ref{fig:SwAV}). Based on SwAV, SEER~\cite{goyal2021self} trained a RegNetY~\cite{radosavovic2020designing} with 1.3B parameters on 1B random images, which for the first time surpassed the best supervised pre-trained model. In addition, some recent works presented other efforts towards bridging contrastive learning and clustering, such as Prototypical Contrastive Learning (PCL)~\cite{li2020prototypical}, Jigsaw clustering~\cite{Chen_2021_CVPR} and Contrastive Clustering (CC) \cite{li2021contrastive}.

\textbf{Remote sensing.} Saha \etal~\cite{saha2021self} proposed to use DeepCluster and triplet contrastive learning to encode multi-modal multi-temporal remote sensing images for change detection. 
\textcolor{blue}{The authors further integrate DeepCluster, BYOL and MoCo-v2 in the pixel-level to  produce segmentation maps \cite{saha2022unsupervised}.} 
Walter \etal~\cite{walter2020self} presented a comparison between DeepCluster, VAE, clolorization as \textcolor{blue}{pretext} task and BiGAN for remote sensing image retrieval. Similarly, Cao \etal~\cite{cao2021contrastnet} presented a comparison between VAE, AAE and PCL for hyperspectral image classification. Hu \etal~\cite{hu2021deep} proposed  spatial-spectral subspace clustering for hyperspectral images based on contrastive clustering \cite{li2021contrastive}. \textcolor{blue}{Liu \etal~\cite{liu2022contrastive} proposed  dual dynamic graph convolutional network (DDGCN), contributing a novel clustering-based contrastive loss to capture the structures of views and scenes. In addition to naive contrastive learning on views, scene structures are added into the loss term by clustering scene indexes within the minibatch.}

\subsubsection{Knowledge Distillation}

Another set of methods for contrastive self-supervised learning is based on knowledge distillation~\cite{hinton2015distilling}. These methods commonly use a teacher-student network (still Siamese-like structure) and optimize a similarity metric of two augmented views of the same input image. Either asymmetric learning rules or asymmetric architectures are utilized to transfer knowledge between the student network and the teacher network, and thus no negative samples are necessary.

\begin{figure}
\centering
\includegraphics[width=0.9\linewidth]{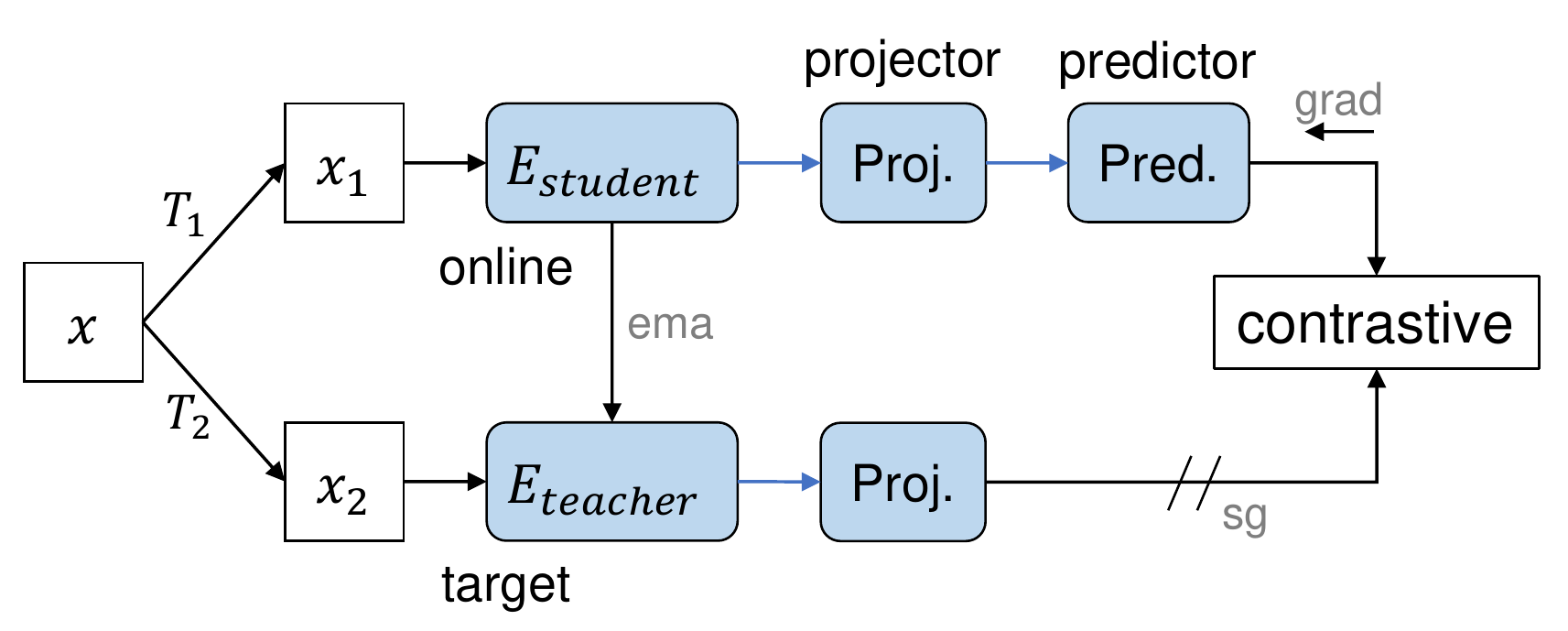}
\caption[BYOL]{Bootstrap Your Own Latent (BYOL)~\cite{grill2020bootstrap}. A similarity loss is minimized between features from the online (student) and the target (teacher) branch. A stop-gradient ($sg$) operator is applied on the teacher whose parameters are updated with an exponential moving average ($ema$) of the student parameters.
}
\label{fig:BYOL}
\end{figure}

Grill \etal~\cite{grill2020bootstrap} proposed a first milestone called BYOL (bootstrap your own latent, see Fig. \ref{fig:BYOL}) for self-supervised learning based on knowledge distillation. The general architecture is similar to MoCo~\cite{he2020momentum} but without the usage of negative samples. It was shown that, if a fixed randomly initialized network (which would not collapse because it is not trained) is used to serve as the key encoder, the representation produced by the query encoder would still be improved during training. If then the target encoder is set to be the trained query encoder and iterates this procedure, it will progressively achieve better performance. Therefore, BYOL proposed an architecture with an exponential moving average strategy to update the target encoder just as MoCo does, and used mean square error as the similarity measurement. Additionally, BYOL is also similar to mean teacher~\cite{tarvainen2017mean} (a semi-supervised method including also a classification loss), which would collapse if removing the classification loss. To prevent collapse, BYOL introduced an additional predictor on top of the online (teacher) network.

SimSiam~\cite{chen2021exploring} presented a systematic study on the importance of different tricks in avoiding collapse and proposed a simplified version of the previous self-supervised contrastive methods, arguing that the additional predictor of BYOL is helpful but not necessary to prevent model collapse. Instead, the stop gradient operation of the teacher (target) network is the most critical component to make target representation stable. In addition, the authors showed the relationship between SimSiam and other popular contrastive methods: (1) SimSiam can be thought of as “SimCLR without negatives”; (2)  SimSiam is conceptually analogous to “SwAV without online clustering”; (3) SimSiam can be seen as "BYOL without the momentum encoder".

DINO~\cite{caron2021emerging} further explored the self-distillation scheme with vision transformer backbones, proposing a new state-of-the-art self-supervised learning algorithm. Following a common teacher-student network architecture, the output of the teacher network is centered with a mean computed over the batch. Each network outputs a K dimensional feature that is normalized with a temperature softmax over the feature dimension and their similarity is then measured with a cross-entropy loss. Stop gradient and momentum update are used to improve the performance. A following work EsViT~\cite{li2021efficient} improved DINO by introducing additional region-level contrastive task apart from commonly used global view-level task. Recently, Zhou \etal~\cite{zhou2021ibot} integrated masked image modeling into the contrastive self-distillation scheme with vision transforms, proposing image BERT pre-training with Online Tokenizer (iBOT).

\begin{figure}
\centering
\includegraphics[width=0.9\linewidth]{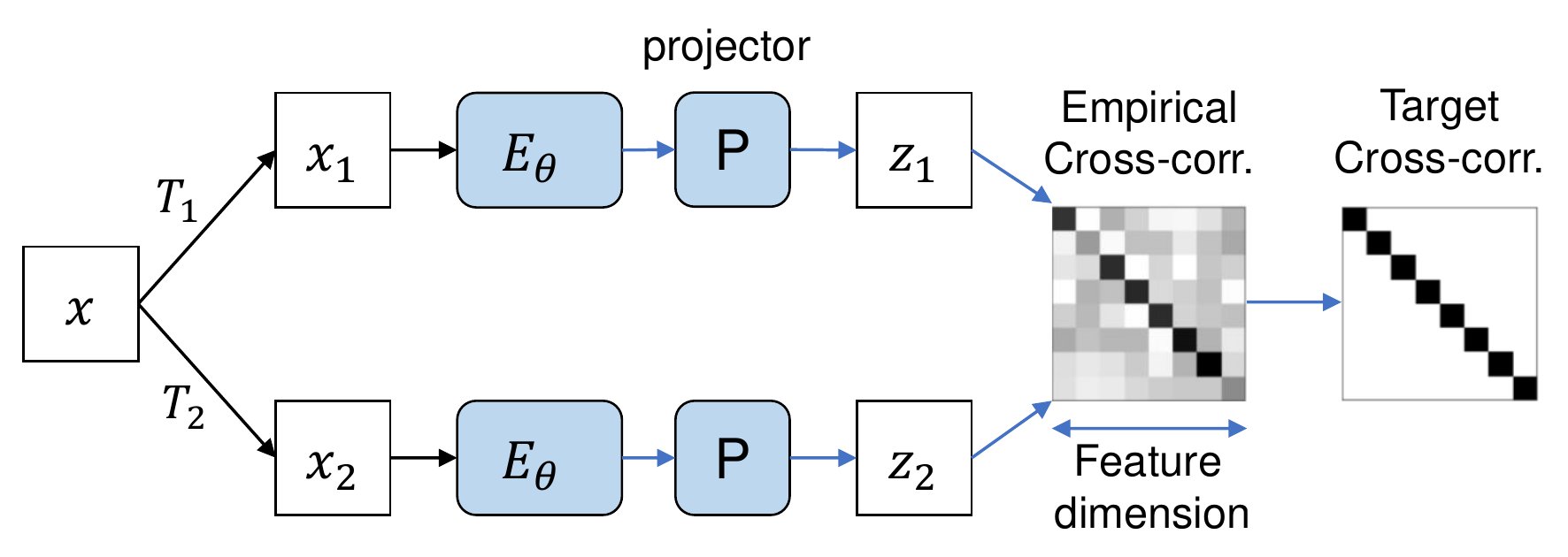}
\caption[Barlow Twins]{Barlow Twins~\cite{zbontar2021barlow}. The objective function measures the crosscorrelation matrix between the embeddings of two identical networks fed with distorted versions of a batch of samples and tries to make this matrix close to the identity. This causes the embedding vectors of distorted versions of a sample to be similar while minimizing the redundancy between the components of these vectors.}
\label{fig:BarlowTwins}
\end{figure}

\begin{figure*}
\centering
\includegraphics[width=0.9\linewidth]{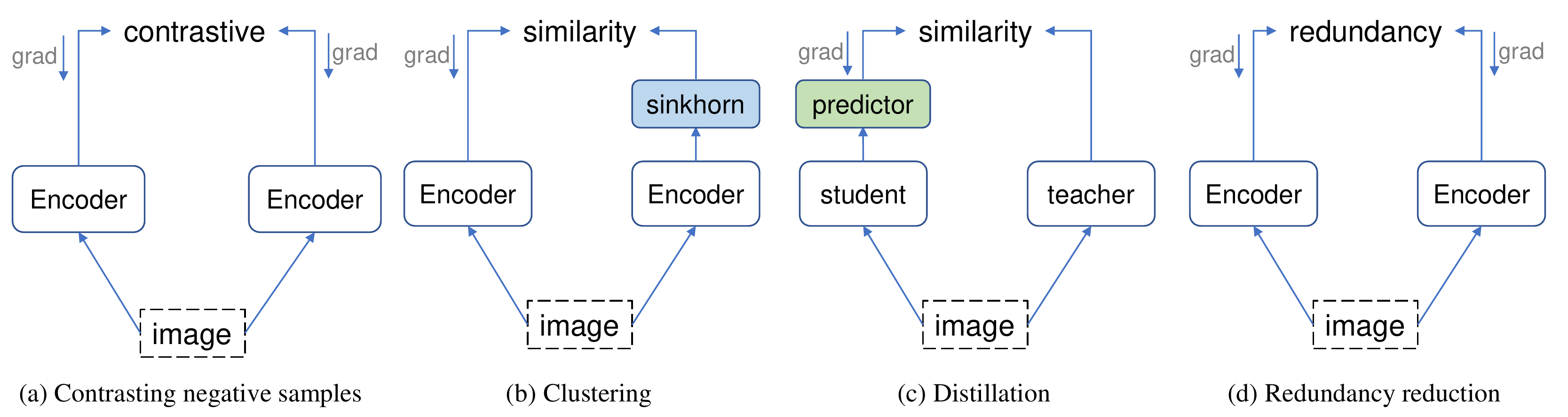}
\caption[taxonomy-contrastive]{A comparison of contrastive self-supervised methods.}
\label{fig:taxonomy-contrastive}
\end{figure*}

\begin{figure*}[h!]
  \begin{minipage}[b]{0.50\linewidth}
    \centering
    \includegraphics[width=\linewidth]{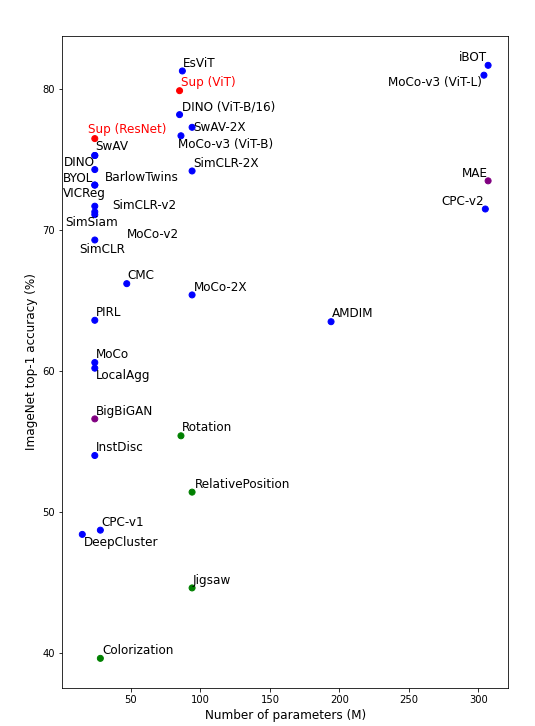}
  \end{minipage}%
  \begin{minipage}[b]{0.50\linewidth}
    \centering
    \scalebox{0.8}{
\begin{tabular}[b]{llcc}
\hline
Method           & Architecture  & Params (M) & Top-1 acc \\ \hline
BigBiGAN~\cite{donahue2019large}         & R50           & 24             & 56.6           \\
MAE~\cite{he2021masked}                  & ViT-L/16         & 307            & 73.5        \\
RelativePosition~\cite{doersch2015unsupervised} & R50w2X        & 94             & 51.4           \\
Jigsaw~\cite{noroozi2016unsupervised}           & R50w2X        & 94             & 44.6           \\
Rotation~\cite{gidaris2018unsupervised}         & Rv50w4X       & 86             & 55.4           \\
Colorization~\cite{zhang2016colorful}     & R101          & 28             & 39.6           \\
CPC-v1~\cite{oord2018representation}           & R101          & 28             & 48.7           \\
CPC-v2~\cite{henaff2020data}          & R161          & 305            & 71.5           \\
AMDIM~\cite{bachman2019learning}           & R-custom & 194            & 63.5           \\
CMC~\cite{tian2020contrastive}             & R50           & 47             & 66.2           \\
InstDisc~\cite{wu2018unsupervised}        & R50           & 24             & 54.0           \\
PIRL~\cite{misra2020self}            & R50           & 24             & 63.6           \\
MoCo~\cite{he2020momentum}            & R50           & 24             & 60.6           \\
MoCo-2X          & R50w2X        & 94             & 65.4           \\
SimCLR~\cite{chen2020simple}          & R50           & 24             & 69.3           \\
SimCLR-2X        & R50w2X        & 94             & 74.2           \\
MoCo-v2~\cite{chen2020improved}         & R50           & 24             & 71.1           \\
SimCLR-v2~\cite{chen2020big}       & R50           & 24             & 71.7           \\
MoCo-v3 (ViT-B)~\cite{chen2021empirical}   & ViT-B/16         & 86             & 76.7           \\
MoCo-v3 (ViT-L)    & ViT-L/16    & 304            & 81.0           \\
DeepCluster~\cite{caron2018deep}     & VGG           & 15             & 48.4           \\
LocalAgg~\cite{zhuang2019local}        & R50           & 24             & 60.2           \\
SwAV~\cite{caron2020unsupervised}            & R50           & 24             & 75.3           \\
SwAV-2X          & R50w2X        & 94             & 77.3           \\
BarlowTwins~\cite{zbontar2021barlow}     & R50           & 24             & 73.2           \\
VICReg~\cite{bardes2021vicreg}          & R50           & 24             & 73.2           \\
BYOL~\cite{grill2020bootstrap}             & R50           & 24             & 74.3           \\
SimSiam~\cite{chen2021exploring}         & R50           & 24             & 71.3           \\
DINO (RN)~\cite{caron2021emerging}            & R50           & 24             & 75.3           \\
DINO (ViT)  & ViT-B/16      & 85             & 78.2           \\
EsViT~\cite{li2021efficient}           & Swin-B        & 87             & 81.3           \\ 
iBOT~\cite{zhou2021ibot}            & ViT-L/16         & 307            & 81.7           \\    
Supervised (ResNet)    & R50    & 24     & 76.5      \\
Supervised (ViT)       & ViT-B/16    & 85    & 79.9    \\ \hline
\end{tabular}
}
\end{minipage}
\caption{Comparison of SSL methods under the linear classification protocol on ImageNet \cite{deng2009imagenet}. All are reported as unsupervised pre-training on the ImageNet-1M training set, followed by supervised linear classification trained on frozen features, evaluated on the validation set. The parameter counts are those of the feature extractors which are commonly ResNet \cite{he2016deep} or ViT \cite{dosovitskiy2020image}.}
\label{fig:imagenet-accuracy}
\end{figure*}

\textbf{Remote sensing.} Guo \etal~\cite{guo2021self} proposed to combine GAN and BYOL for better discriminative representation learning. Following a normal GAN structure, a generator is used to generate fake images and a discriminator is used to distinguish them from real images. An additional similarity loss is combined with the discriminator loss by seeing the discriminator as a self-supervised encoder, which encodes both fake and real images as two input views to a BYOL-like Siamese network. In addition, gated self-attention and pyramidal convolution are proposed to be used in both the generator and the discriminator. Chen \etal~\cite{chen2021self0,chen2021self2} utilized Siamese ResUNet to distill related representations from different modalities (SAR and optical), and a shift transformation to learn pixel-wise feature representations. Hu \etal~\cite{hu2021contrastive} utilized transformer as encoder backbone and BYOL as baseline structure for hyperspectral image classification. 
\textcolor{blue}{
Dong \etal~\cite{dong2021exploring} employed SimSiam for ViT based PolSAR image classification. Wang \etal~\cite{wang2022fiad} leverage BYOL for SAR ship detection. Jain \etal~\cite{jain2022self} perform pre-training by contrasting SAR and optical images using BYOL. Wang \etal~\cite{wang2022self} proposed DINO-MM, a joint SAR-optical representation learning approach with vision transformers. Following the main self-supervised mechanism as DINO, the authors introduced RandomSensorDrop to let the model see all possible combinations of both modalities during training. Zhang \etal~\cite{zhang2022self} introduced a self-supervised spectral–spatial attention-based vision transformer (SSVT), where global and local augmented views are contrasted based on self-distillation \cite{zhang2019your}. Muhtar \etal~\cite{muhtar2022index} proposed IndexNet, a dense self-supervised method for remote sensing image segmentation. Their approach is built on BYOL and performs contrastive at image and pixel level to preserve spatial information. In a similar fashion, Chen \etal~\cite{chen2022semantic} propose a pixel-level self supervised learning approach for change detection. Their method is based on SimSiam with the objective of enforcing point-level consistency across views. They also propose a background-swap augmentation to focus on the foreground.
}

\subsubsection{Redundancy Reduction}

The idea of redundancy reduction for self-supervised learning comes from neuroscience. Indeed, H. Barlow~\cite{barlow1961possible} states that the goal of sensory processing is to record highly redundant sensory inputs into a factorial code (a code with statistically independent components). Inspired by this principle, Zbontar \etal~ proposed Barlow Twins~\cite{zbontar2021barlow}, a method that uses redundancy reduction as a way to avoid trivial solution in contrastive learning without explicit (e.g. MoCo~\cite{he2020momentum} and SimCLR~\cite{chen2020simple})  or implicit (e.g. DeepCluster~\cite{caron2018deep} and SwAV~\cite{caron2020unsupervised}) negative samples. With the main network architecture similar to previous contrastive learning, Barlow Twins’ objective function measures the cross-correlation matrix between the embeddings of two identical networks fed with distorted versions of a batch of samples, and tries to make this matrix close to the identity (Fig. \ref{fig:BarlowTwins}). This causes the embedding vectors of distorted versions of a sample to be similar while minimizing the redundancy between the components of these vectors. The loss function of Barlow Twins is:

\begin{equation}
\mathcal{L}_{\mathcal{B} \mathcal{T}} \triangleq \underbrace{\sum_{i}\left(1-\mathcal{C}_{i i}\right)^{2}}_{\text {invariance term }}+\lambda \underbrace{\sum_{i} \sum_{j \neq i} \mathcal{C}_{i j}^{2}}_{\text {redundancy reduction term }}
\end{equation}

\noindent where $\lambda$ is a trade-off parameter and $C$ is the cross-correlation matrix computed between the outputs of the two identical networks along the batch dimension:

\begin{equation}
\mathcal{C}_{i j} \triangleq \frac{\sum_{b} z_{b, i}^{A} z_{b, j}^{B}}{\sqrt{\sum_{b}\left(z_{b, i}^{A}\right)^{2}} \sqrt{\sum_{b}\left(z_{b, j}^{B}\right)^{2}}}
\end{equation}

\noindent where $b$ indexes batch samples, $i$, $j$ index the vector dimension of the networks’ outputs, and $A$, $B$ represent two different views.

Bardes \etal~\cite{bardes2021vicreg} borrowed the decorrelation mechanism from Barlow Twins, proposing a new redundancy-reduction-based self-supervised method VICReg (variance-invariance-covariance regularization). VICReg follows common contrastive learning architecture, and avoids the trivial solution by introducing variance along the batch dimension (a hinge loss which constrains the standard deviation), invariance cross different views (mean-squared euclidean distance), and covariance along feature dimension (penalizing the off-diagonal coefficients of the covariance matrix of the embeddings).

\textbf{Remote sensing.} Since redundancy-reduction-based algorithms (Barlow Twins \cite{zbontar2021barlow} and VICReg \cite{bardes2021vicreg}) are relatively new, there are only works using these methods in the remote sensing community. \textcolor{blue}{Marsocci \etal~\cite{marsocci2022continual} proposed to use Barlow-Twins in a continual learning setting to pre-train a model on a set of heterogeneous datasets while avoiding catastrophic forgetting. Their approach is evaluated on segmentation problems. Ait Ali Braham \etal~ \cite{nassim2022self} used Barlow-Twins for few-shot HSI classification in a two-stages approach. They propose two pair sampling strategies for pixel and patch level classification using spatial information from the scene to generate positive pairs, along with a set of data augmentations. Their results support the effectiveness of redundancy reduction methods on hyperspectral images.} In the next section, we further discuss the conceptual idea of redundancy reduction and how it could be beneficial for remote sensing image understanding. We believe there is a large potential to be explored.

\section{From ImageNet to remote sensing imagery: self-supervised learning on geospatial earth observation data}
\label{sec:ssl-in-rs}

Self-supervised learning is a data-driven methodology that learns the representation of a dataset without human annotation. However, most existing self-supervised methods in the computer vision community deal with natural RGB images (e.g., ImageNet~\cite{deng2009imagenet}) which are different from remote sensing imagery, and there are various types of data from various types of sensors that require specific considerations. In addition, though intended to learn common representations for various downstream tasks, there's an unignorable gap between pixel-level, patch-level and image-level tasks that may benefit from different designs of self-supervision. In this section, we dig into the implementation of self-supervised methods on different types of remote sensing data and image-/patch-/pixel-level applications.

\begin{table*}
\caption{A gallery of recent self-supervised works in remote sensing.}
\label{tab:ssl-rs}
\centering
\scalebox{1.}{

\begin{tabular}{|c|c|l|}
\hline
Modality                         & Application                                   & \multicolumn{1}{c|}{Method}                                                                     \\ \hline
\multirow{27}{*}{Multi-spectral} & \multirow{10}{*}{Representation learning}     & Vincenzi \etal~\cite{vincenzi2021color}: colorization as pretext task.                          \\
                                 &                                               & tile2vec~\cite{jean2019tile2vec}: contrastive learning with triplet loss.                       \\
                                 &                                               & Jung \etal~\cite{jung2021self}: replace triple loss of tile2vec to binary cross entropy loss.   \\
                                 &                                               & Stojnic \etal~\cite{stojnic2021self}: contrastive multiview coding.                             \\
                                 &                                               & CSF~\cite{swope2021representation}: contrastive multiview coding.                               \\
                                 &                                               & Jung \etal~\cite{jung2021contrastive}: SimCLR with smoothed view.                               \\
                                 &                                               & GeoMoCo~\cite{ayush2020geography}: MoCo + geo-location as pretext task.                         \\
                                 &                                               & SauMoCo~\cite{kang2020deep}: MoCo with spatial augmentation.                                    \\
                                 &                                               & SeCo~\cite{manas2021seasonal}: MoCo + seasonal contrast.                                        \\
                                 &                                               & GeoKR~\cite{li2021geographical}: geo-supervision (landcover map) + teacher-student network.     \\ \cline{2-3} 
                                 & \multirow{4}{*}{Scene classification}         & Lu \etal~\cite{lu2017remote}: autoencoder.                                                      \\
                                 &                                               & Zhao \etal~\cite{zhao2020self}: rotation as pretext task + classification loss.                 \\
                                 &                                               & Tao \etal~\cite{tao2020remote}: inpainting, relative position and instance discrimination.      \\
                                 &                                               & SGSAGANs~\cite{guo2021self}: BYOL + GAN.                                                        \\ \cline{2-3} 
                                 & \multirow{3}{*}{Semantic segmentation}        & Li \etal~\cite{li2021semantic}: inpainting/rotation as pretext tasks + contrastive learning.    \\
                                 &                                               & Singh \etal~\cite{singh2018self}: GAN + inpainting as pretext task.                             \\
                                 &                                               & Li \etal~\cite{li2021remote}: global and local contrastive learning.                            \\ \cline{2-3} 
                                 & \multirow{6}{*}{Change detection}             & Zhang \etal~\cite{zhang2016change}: denoising Autoencoder.                                      \\
                                 &                                               & S2-cGAN~\cite{alvarez2020s2}: conditional GAN discriminator + reconstruction loss.              \\
                                 &                                               & Dong \etal~\cite{dong2020self}: GAN discriminator for temporal prediction.                      \\
                                 &                                               & Cai \etal~\cite{cai2021task}: clustering for hard sample mining.                                \\
                                 &                                               & Leenstra \etal~\cite{leenstra2021self}: triplet loss + binary cross entropy loss.               \\
                                 &                                               & Chen \etal~\cite{chen2021self1}: contrastive multiview coding + BYOL.                           \\ \cline{2-3} 
                                 & Time series classification                    & Yuan \etal~\cite{yuan2020self}: transformer + temporal pretext task.                            \\ \cline{2-3} 
                                 & Object detection                              & DUPR~\cite{ding2021unsupervised}: patch re-identification + multi-level contrastive loss        \\ \cline{2-3} 
                                 & Image retrieval                               & Walter \etal~\cite{walter2020self}: DeepCluster/BiGAN/VAE/Colorization.                         \\ \cline{2-3} 
                                 & Depth estimation                              & Madhuanand \etal~\cite{madhuanand2021self}: autoencoder + multi-task prediction + triplet loss. \\ \hline
\multirow{11}{*}{Hyperspectral}  & \multirow{6}{*}{Image classification}         & Mou \etal~\cite{mou2017unsupervised}: autoencoder.                                              \\
                                 &                                               & Li \etal~\cite{li2021self}: autoencoder + subspace clustering.                                  \\
                                 &                                               & Liu \etal~\cite{liu2020deep}: contrastive multi-view coding.                                    \\
                                 &                                               & Hu \etal~\cite{hu2021deep}: spatial-spectral contrastive clustering.                            \\
                                 &                                               & Cao \etal~\cite{cao2021contrastnet}: VAE/AAE/PCL.                                               \\
                                 &                                               & Hu \etal~\cite{hu2021contrastive}: BYOL + transformer.                                          \\ \cline{2-3} 
                                 & Unmixing                                      & EGU-Net~\cite{hong2021endmember}: autoencoder + Siamese weight-sharing network.                 \\ \cline{2-3} 
                                 & Target detection                              & Yao \etal~\cite{yao2021self}: clustering + pseudo-label + contrastive N-pair loss.              \\ \cline{2-3} 
                                 & Denoising                                     & SHDN~\cite{wang2021self}: noise prediction as pretext task.                                     \\ \cline{2-3} 
                                 & Restoration                                   & Imamura \etal~\cite{imamura2019self}: denoising autoencoder.                                    \\ \cline{2-3} 
                                 & Super-resolution                              & CUCaNet~\cite{yao2020cross}: autoencoder + Siamese cross-attention.                             \\ \hline
\multirow{7}{*}{SAR}             & \multirow{3}{*}{Despeckling}                  & BDSS~\cite{yuan2019blind}: autoencoder, predict another speckled view.                          \\
                                 &                                               & Speckle2Void~\cite{molini2021speckle2void}: blind-spot CNN + physical regression.               \\
                                 &                                               & Wang \etal~\cite{wang2021sar}: Siamese contrastive loss + pseudo label.                         \\ \cline{2-3} 
                                 & Scene classification                          & MI-SSL~\cite{ren2021mutual}: PolSAR modal similarity + instance contrastive.                    \\ \cline{2-3} 
                                 & Image retrieval                               & Park \etal~\cite{park2021homography}: MoCo.                                                     \\ \cline{2-3} 
                                 & \multirow{2}{*}{Target recognition}           & RR-SSL~\cite{zhang2019rotation}: rotation as pretext task.                                      \\
                                 &                                               & UACL~\cite{xu2021adversarial}: adversarial contrastive learning.                                \\ \hline
\multirow{8}{*}{multi-modal}     & \multirow{3}{*}{SAR-optical fusion}           & Cha \etal~\cite{cha2021contrastive}: contrastive multiview coding.                              \\
                                 &                                               & UniFeat\&CoRe~\cite{montanaro2021self}: SimCLR + inpainting as pretext.                         \\
                                 &                                               & Chen \etal~\cite{chen2021self0}: instance discrimination + multiview \& shift invariance.       \\ \cline{2-3} 
                                 & \multirow{3}{*}{SAR-optical change detection} & Chen \etal~\cite{chen2021self1}: contrastive multiview coding + BYOL.                           \\
                                 &                                               & Chen \etal~\cite{chen2021self2}: instance discrimination + multiview \& shift invariance.       \\
                                 &                                               & Saha \etal~\cite{saha2021self}: DeepCluster + bitemporal multi-sensor contrastive.              \\ \cline{2-3} 
                                 & SAR-optical matching                          & Hughes \etal~\cite{hughes2018mining}: VAE+GAN, hard negative sample mining.                     \\ \cline{2-3} 
                                 & Audio-image fusion                            & Heidler \etal~\cite{heidler2021self}: audio-visual contrastive triplet loss.                    \\ \hline
\end{tabular}

}
\end{table*}

\begin{figure}
\centering
\includegraphics[width=0.9\linewidth]{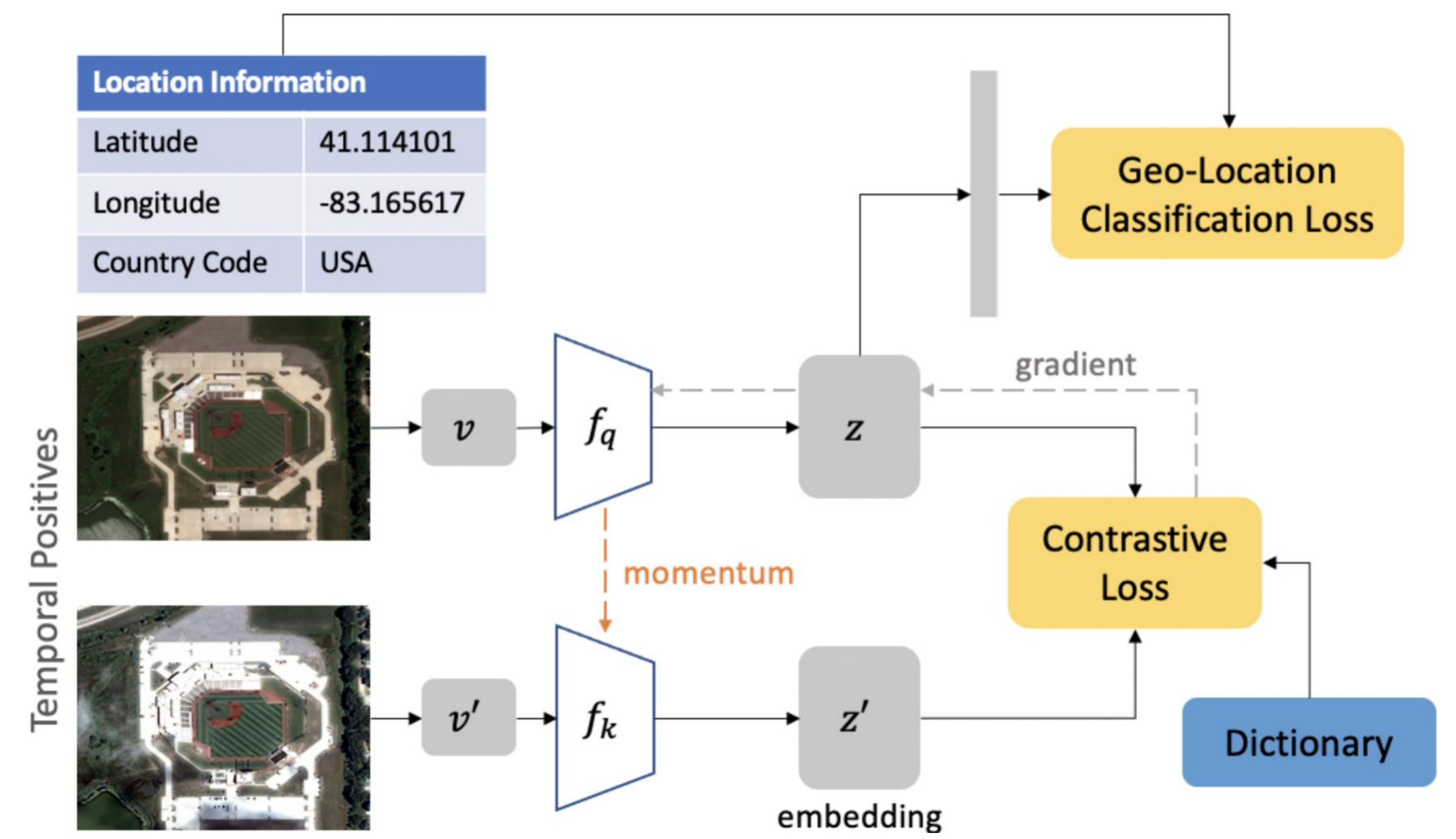}
\caption[geomoco]{Geography-aware self-supervised learning~\cite{ayush2020geography}. Spatially aligned images over time are used to construct temporal positive pairs in contrastive learning, and geo-location is used as an additional pre-text task to boost representation learning for remote sensing images. ©[2021] IEEE.}
\label{fig:geomoco}
\end{figure}

\begin{figure*}
\centering
\includegraphics[width=0.9\linewidth]{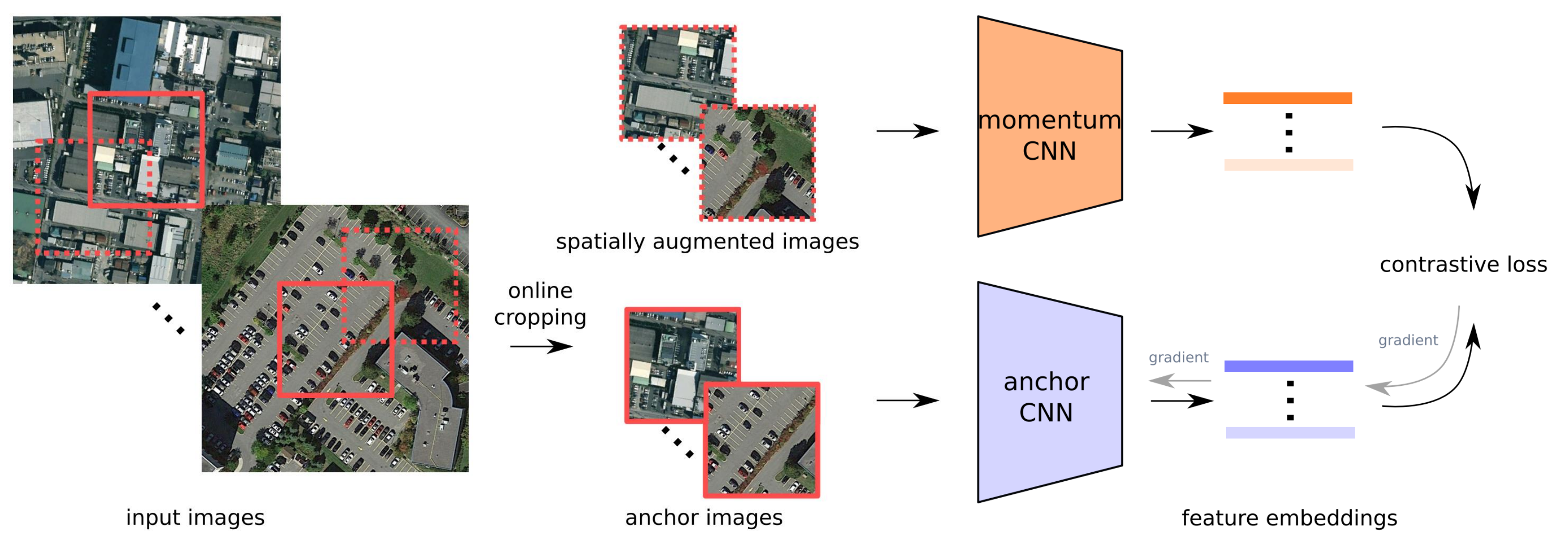}
\caption[saumoco]{Spatially augmented momentum contrast~\cite{kang2020deep}. Following the underlying assumption of Tile2Vec~\cite{jean2019tile2vec}, nearby cropped tiles are seen as spatial augmentation for the anchor tile, and a similarity loss is optimized based on momentum contrast~\cite{he2020momentum}. ©[2020] IEEE.}
\label{fig:saumoco}
\end{figure*}

\subsection{Characterisics and Challenges of Remote Sensing Data}

Remote-sensing data contains multiple modalities, e.g., from optical (multi- and hyperspectral), Lidar, and synthetic aperture radar (SAR) sensors, where the imaging geometries and contents are completely different \cite{zhu2017deep}. Yet before we discuss those different modalities, there exist some characteristics that all data types share as a common remote sensing specific property compared to natural images. 

First, remote sensing data are geolocated, which means each image pixel corresponds to a geospatial coordinate. This information provides additional chances for the design of self-supervised pretext tasks (e.g., predicting the geo-location of an input image~\cite{ayush2020geography}, see Fig. \ref{fig:geomoco}), as well as indirect cooperation with geographical resources (e.g., land cover database~\cite{li2021geographical}).

Second, remote sensing data are geodetic measurements and usually correspond to specific physical meanings. If not well analyzed, this characteristic can raise issues with data augmentation. Specifically, data augmentation is very important especially for contrastive methods to help the network learn semantic invariance, yet a careless design of data augmentation might change certain physical properties which are indeed important to recognize the objects in a remote sensing image.

Third, remote sensing images usually contain many more objects than natural images (e.g., compared to an image from ImageNet which generally contains one main object, a remote sensing image usually includes multiple repeated objects). This difference influences both pretext tasks and data augmentation. For example, jigsaw puzzles can be expected to be difficult for representation learning on low-resolution satellite images, and rotation can affect the shadows of buildings which are important for the model to learn relative heights. On the other hand, when image-level representation is important for natural images, patch-level and pixel-level representation can be equally important for remote sensing images (e.g., considering the spatial relationship of neighboring and distant patches to sample positive/negative pairs~\cite{jean2019tile2vec,kang2020deep} as shown in Fig. \ref{fig:tile2vec} and \ref{fig:saumoco}). 

\begin{figure}
\centering
\includegraphics[width=0.9\linewidth]{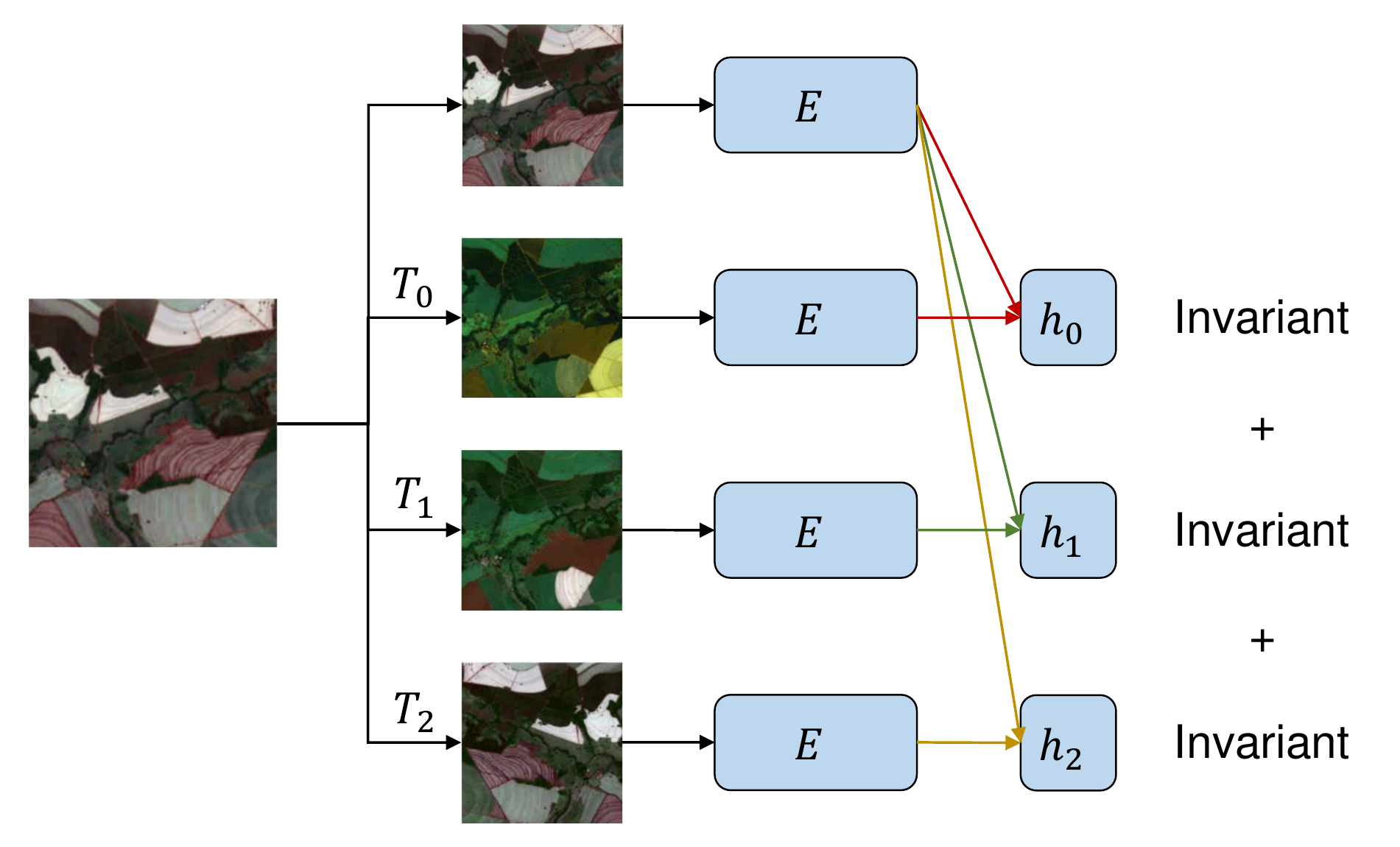}
\caption[seco]{Seasonal Contrast (SeCo)~\cite{manas2021seasonal}. Three embedding subspaces are trained to be invariant to temporal (seasonal), synthetic and both temporal and synthetic transforms $T_{1}$, $T_{2}$ and $T_{0}$, respectively.}
\label{fig:seco}
\end{figure}

Last but not least, the time variable is becoming increasingly important despite the modality. This raises the interest of getting inspiration from video self-supervised learning, yet time stamps in remote sensing images are quite different from frames in videos, as well as the special objectives (e.g. change detection in remote sensing). Therefore, how to make the best of temporal information is a common question to be answered in all modalities. An early progress is shown in Fig. \ref{fig:seco} where seasonal information is used for data augmentation \cite{manas2021seasonal}.

Apart from these important common characteristics, below we discuss different modalities of remote sensing data. A list of recent self-supervised works classified w.r.t modality is shown in Table~\ref{tab:ssl-rs}.

\subsubsection{Multispectral Images}

\begin{figure}
\centering
\includegraphics[width=0.95\linewidth]{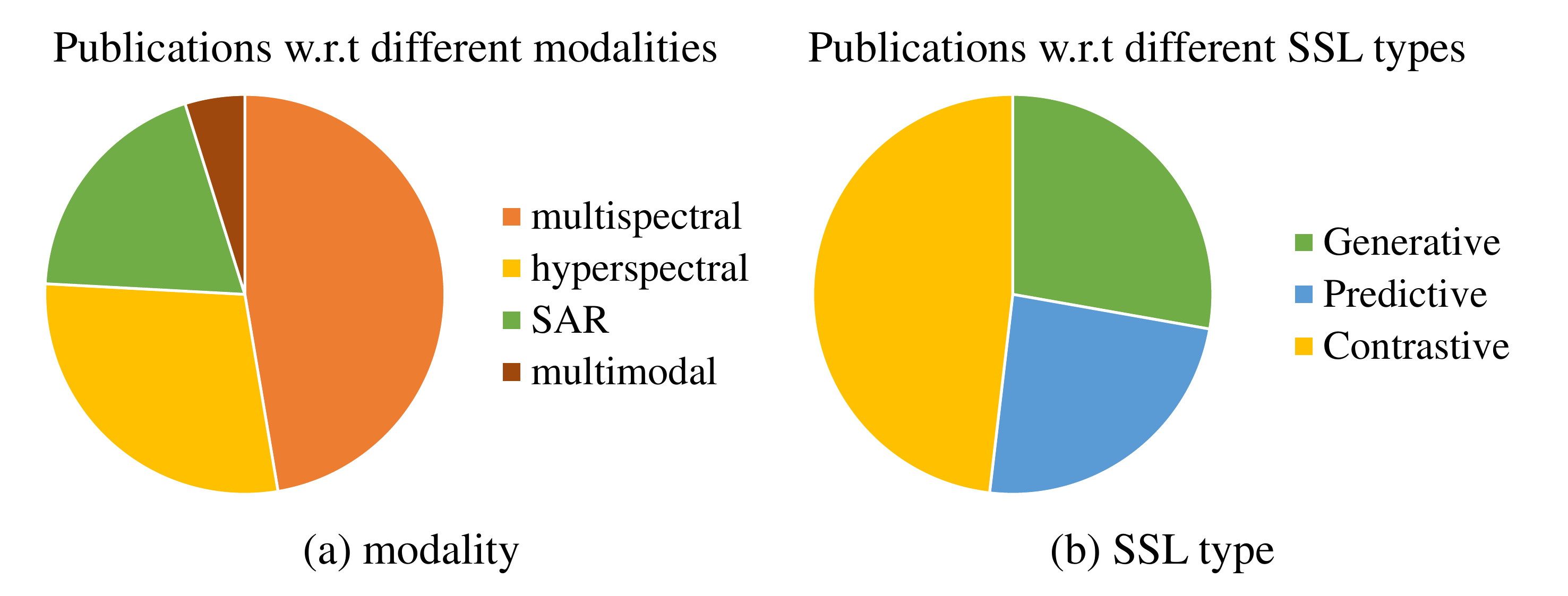}
\caption[publications ssl in rs]{Statistics of recent publications related to self-supervised learning in remote sensing.}
\label{fig:ssl-publications}
\end{figure}

\textcolor{blue}{Multispectral images are captured within specific wavelengths ranging across the electromagnetic spectrum, which allows for the extraction of information covering or beyond the perception range of human eyes.} Compared to hyperspectral imaging, multispectral imaging measures light in a small number (typically 3 to 15) of spectral bands with relatively low spectral resolution but high spatial resolution. Multispectral images are the most commonly used remote sensing data, as they are relatively easy to acquire, and are close to natural images which enable easier human perception and convenient technology transfer from the computer vision community. Due to the popularity and the similarity of multispectral images to natural images, a lot of self-supervised methods have been directly transferred to the remote sensing field (see Fig. \ref{fig:ssl-publications}). Generally, all categories of self-supervised methods in CV can be extended to remote sensing multi-spectral image analysis, but there is some special attention to pay: (1) the common characteristics of remote sensing data for all modalities need to be considered as have been discussed above, (2) spectral contexts of predictive methods (e.g., colorization) now have possibly more channel information that can be used to design the pretext tasks, (3) color-related data augmentations (e.g., color distortions shown in Fig. \ref{fig:dataaug_simclr}) which are very important for contrastive self-supervised learning need to be carefully modified to fit into multiple channels.

\subsubsection{Hyperspectral Images}

Hyperspectral imaging also captures images across the electromagnetic spectrum, but lights in a far bigger number (up to hundreds) of spectral bands than multispectral imaging with a trade-off on the spatial resolution. This very high spectral resolution enables us to identify the materials contained in the pixel via spectroscopic analysis, yet, on the other hand, requires special care about band-wise analysis. In fact, with a basic objective of dimensionality reduction, PCA/ICA/clustering and manifold-learning-based methods have been widely studied in the field before the era of neural networks. Along the lines, deep-learning-based self-supervised methods have been focusing on autoencoders and 3D convolution to compress spectral representations. Because of the large number of spectral channels, the network usually takes a single-pixel or a small patch as input. In both cases, spatial contexts for pretext tasks or data augmentations tend to consider intra-sample instead of inner-sample information (e.g., neighboring patches or pixels can be seen as positive samples in contrastive negative sampling). \textcolor{blue}{Meanwhile, spectral contexts have a large potential to be explored because of the high spectral dimensionality of hyperspectral data.} For example, two spectrally divided parts of a hyperspectral image can be two augmented views.

\subsubsection{SAR Images}
Synthetic-aperture radar (SAR) is a form of radar that is used to create two-dimensional images or three-dimensional reconstructions of objects. Due to the imaging principle, SAR images are substantially different from multispectral and hyperspectral images in characteristics like dynamic range, speckle statistics, imaging geometry and complex domain information. While self-supervised learning with SAR images is also a new research direction and has not been widely explored, there's one fact that makes it a valuable topic: the prevailing lack of ground truth for regression-type tasks. Though simulators can be used to provide training data for supervised learning, this bears the risk that the networks will learn models that are far too simplified. Therefore, it would be very beneficial if we can use self-supervised techniques to learn representations from real SAR data. In addition, the general self-supervised schemes can also be easily transferred to SAR images with specific considerations. \textcolor{blue}{For example, the data augmentation of color distortion is not suitable for SAR data, the speckle noise of SAR images has a similar effect as Gaussian blurring, and the ascending and descending order may serve as a natural augmentation strategy.}

\begin{figure}
\centering
\includegraphics[width=0.9\linewidth]{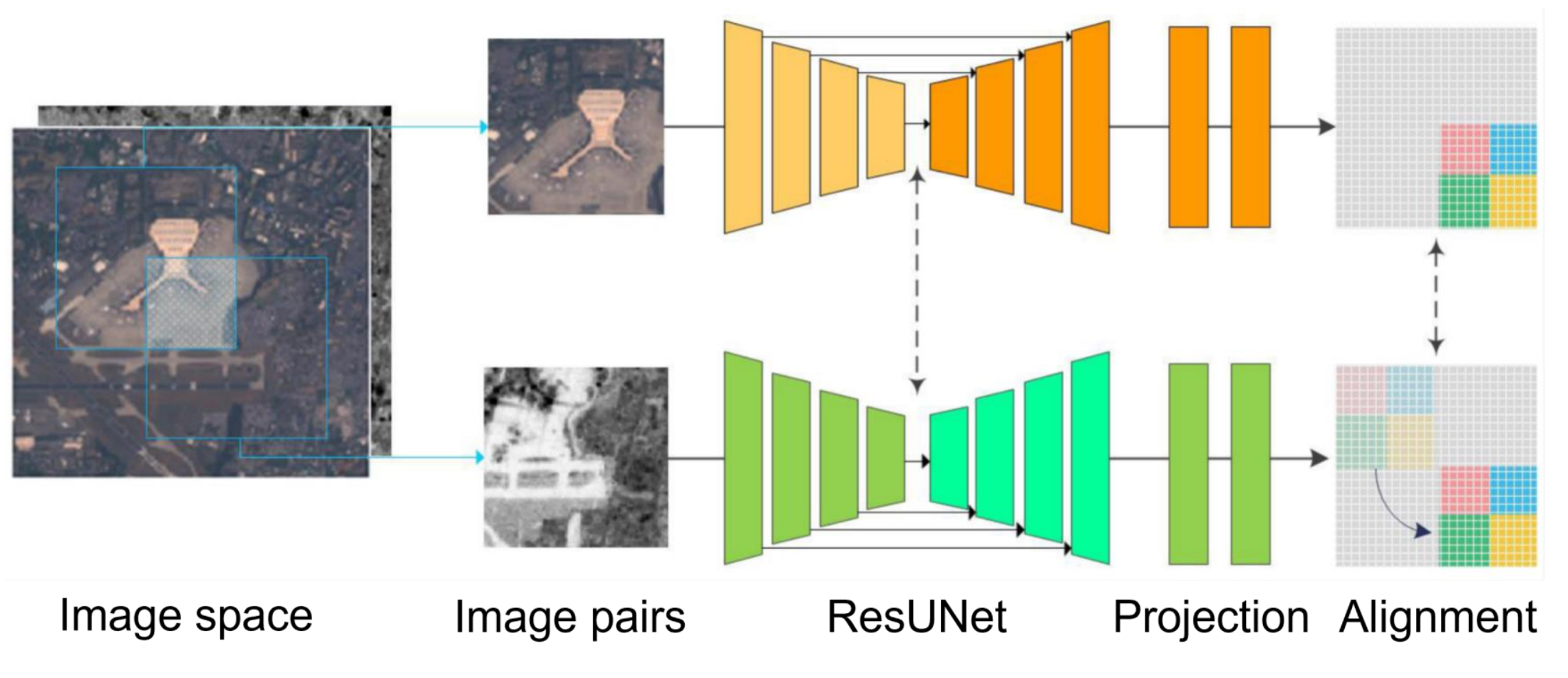}
\caption[PixEF]{Shift invariance~\cite{chen2021self0}. The two inputs have an offset but keep an overlap. A shift transformation is included in the one branch
for aligning representations between two branches which are optimized by a contrastive loss.}
\label{fig:pixef}
\end{figure}

\subsubsection{Multi-sensor Fusion}

As a result of the multi-modality of remote sensing data, multi-sensor fusion is always an important task for various objectives, which can benefit from multi-modal self-supervised learning. There are generally two ways of combining multi-modal inputs: (1) similar to various input channels, different modalities can be seen as augmented views of semantically the same input which can be used for designing pretext tasks or specific data augmentation in contrastive self-supervised learning; (2) two encoders are used separately to encode features from two different modalities and then concatenated together for a joint representation learning. Regarding remote sensing data, a common multi-modality is from optical and SAR images, leading to the important topic SAR-optical data fusion, which has been shown by some recent works that benefits from self-supervised representation learning \cite{chen2021self0} (see Fig. \ref{fig:pixef}) and \cite{wang2022self}. In addition, the fusion of other modalities like audio and social media data is also raising increasing interest \cite{heidler2021self}.

\subsection{Applications}

Though the general goal of self-supervised learning is to perform task-independent representation learning which can benefit various downstream tasks, it has to be noted that different self-supervision can have different influences on different tasks. Therefore, it is necessary to pre-consider the choice of self-supervised methods when we are targeting a specific application in the end. In fact, so far a large number of self-supervised works in the remote sensing field are based on a specific application. In general, the various applications can be separated into three categories: \textit{image-level tasks}, \textit{pixel-level tasks} and \textit{patch-level tasks}. A list of representative recent self-supervised works classified w.r.t applications is shown in Table~\ref{tab:ssl-rs}. \textcolor{blue}{We refer the interested reader to the numerous surveys on machine learning and deep learning in remote sensing \cite{zhu2017deep, ma2019deep, li2018deep, yuan2021review, khelifi2020deep, signoroni2019deep, zhu2021deep} for more details about common downstream tasks and available datasets.}

\subsubsection{Image-level Tasks}

Image-level tasks correspond to the applications that expect the recognition of the whole image or image patches, focusing more on global knowledge. The most common image-level task is scene classification for multispectral \cite{lu2017remote,zhao2020self,tao2020remote,guo2021self}, SAR \cite{ren2021mutual} and other possible modalities. Scene classification is directly related to natural image classification and is usually the default downstream task for the evaluation of a self-supervised method. Like most of the above-mentioned predictive and contrastive self-supervised methods, an image-level task requires focusing on a global representation that can be done by predicting high-level pretexts or contrasting the encoded features of two augmented image views. Other image-level tasks include time series classification \cite{yuan2020self} and image retrieval \cite{walter2020self}.

\subsubsection{Pixel-level Tasks}

Pixel-level tasks correspond to the applications that expect the recognition of each image pixel, focusing more on the local details. The most common pixel-level task is semantic segmentation \cite{Akiva_2021_WACV,li2021semantic,singh2018self,li2021remote}, which requires finer representation in pixel level. Generative self-supervised methods like autoencoder and pretext tasks / data augmentations like image inpainting and pixel based contrastive learning can be helpful. Change detection \cite{zhang2016change,alvarez2020s2,dong2020self,tomenotti2020heterogeneous,cai2021task,chen2021self1,chen2021self2,leenstra2021self,leenstra2021self} is usually also a pixel-level task which utilizes multitemporal information to detect changing pixels. In hyperspectral image analysis, most of the tasks are based on pixel level, including hyperspectral image classification\footnote{Note that with the goal to classify the center pixel, hyperspectral image classification tends to work on a small patch to benefit from spatial relationships, linking towards image and patch level tasks.}\cite{li2021self,kemker2017self,mou2017unsupervised}, image denoising \cite{wang2021self}, spectral unmixing \cite{palsson2022blind}, target detection \cite{yao2021self}, image restoration \cite{imamura2019self} and super-resolution \cite{chen2021hyperspectral,yao2020cross}. Other pixel-level tasks include depth estimation \cite{hermann2020self,madhuanand2021self} and SAR despeckling \cite{yuan2019blind,molini2021speckle2void}.

\subsubsection{Patch-level Tasks}

Patch-level tasks mainly correspond to object detection, lying in-between image-level and pixel-level applications. When requiring both the location and the label of the object in the image, patch-level tasks benefit from both global representation and local details. Hence for good performance in the application of object detection, an integrated representation learning pipeline deserves consideration  \cite{ding2021unsupervised}. \textcolor{blue}{In addition, patch-based change detection considers comparisons between tiles of images (change or no change) rather than pixels due to reasons like possible misalignment between time stamps \cite{saha2021patch}, or the requirement of fast pre-processing to retrieve region of interest and limited bandwidth for data communication \cite{ruuvzivcka2021unsupervised}. We note that patch-level change detection can also be classified into image-level tasks. However, we put them here as there is a stronger correlation between the small patches and the big scene.}

\begin{figure*}
\centering
\subfigure[]{\includegraphics[width=0.55\linewidth]{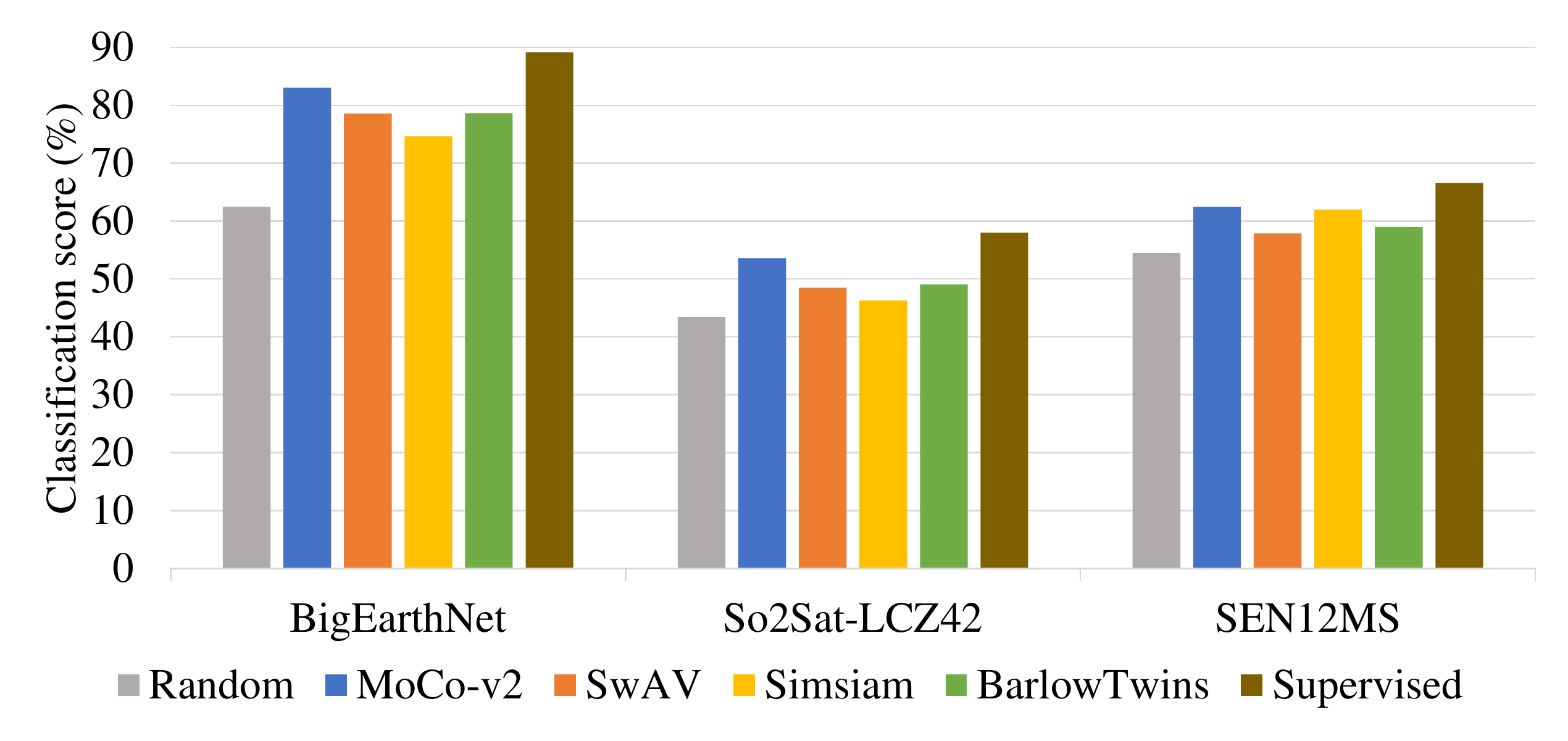}}
\subfigure[]{\includegraphics[width=0.35\linewidth]{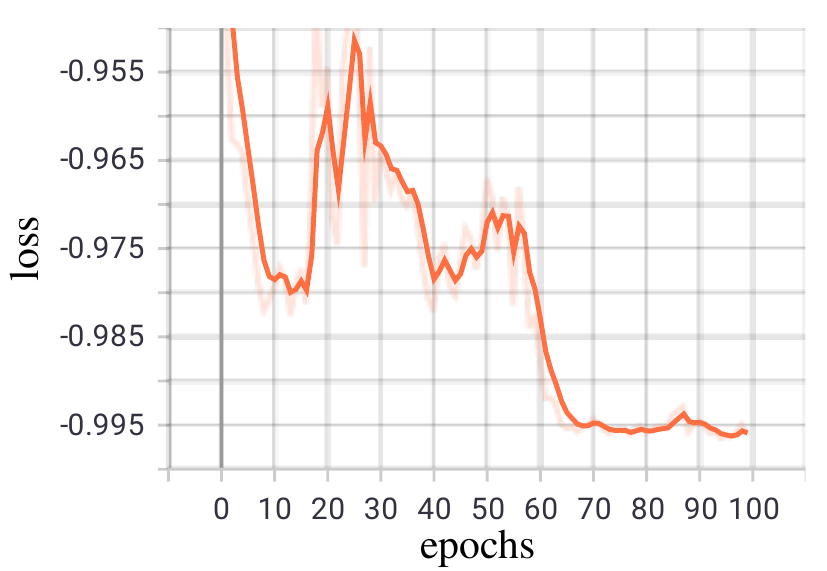}}
\caption[benchmark-SSL-RS]{$(a)$ A preliminary benchmark of four popular contrastive self-supervised methods (each representing one category in section \ref{sec:ContrastiveMethods}) on Sentinel-2 images of BigEarthNet~\cite{sumbul2019bigearthnet}, SEN12MS~\cite{schmitt2019sen12ms} 
and So2Sat-LCZ42~\cite{zhu2019so2sat} (culture-10 version) datasets. We use ResNet-18 as encoder backbones, follow the official settings of the models for self-supervised pre-training, and evaluate the performance by training a linear classifier on frozen features. 
\textcolor{blue}{
$(b)$ The "collapsing" training curve of SimSiam. The loss goes down quickly in the beginning 10 epochs and becomes very unstable in the following epochs.
}
}
\label{fig:ssl-benchmark}
\end{figure*}

\color{blue}
\section{A Preliminary Benchmark}

\label{sec:benchmark-ssl-rs}

As a milestone along the road of large-scale unsupervised pre-training in the image domain, the past two years have witnessed big advances with contrastive self-supervised learning. However, as most of the recent representative methods are developed and verified on natural images, there lacks a comprehensive study about their performance on remote sensing imagery. Therefore, with the hope to build a reference for the remote sensing community, we provide in this section a preliminary benchmark of recent contrastive self-supervised methods on several popular remote sensing datasets. Codes are available at \url{https://github.com/zhu-xlab/SSL4EO-Review}.

\subsection{Methods, Datasets and Implementation Details}

\subsubsection{Methods and Datasets} We consider four representative contrastive self-supervised methods to provide the benchmark: MoCo-v2~\cite{chen2020improved}, SwAV~\cite{caron2020unsupervised}, SimSiam~\cite{chen2021exploring} and Barlow Twins~\cite{zbontar2021barlow}, each representing one sub-category of modern contrastive self-supervised learning: negative sampling, clustering, knowledge distillation and redundancy reduction. We use three popular remote sensing datasets for pre-training and evaluation: BigEarthNet~\cite{sumbul2019bigearthnet}, SEN12MS~\cite{schmitt2019sen12ms} and So2Sat-LCZ42~\cite{zhu2019so2sat}. BigEarthNet is a multi-label scene classification dataset, containing 590,326 non-overlapping Sentinel-2 image patches of size 128*128 covering 10 European countries. SEN12MS is a global dataset with 180,662 256*256 triplets of Sentinel-1, Sentinel-2 and MODIS land cover maps. So2Sat-LCZ42 is a dataset built for local climate zone classification, containing 400,673 32*32 Sentinel-1/2 pairs covering 42 cities in the world. They are all large-scale datasets with different geographical coverage and built for different tasks, which we believe provide solid diversity for evaluation of the generalizability of self-supervised methods. In addition, we use EuroSAT~\cite{helber2019eurosat} for transfer learning evaluation, which is a small single-label land cover classification dataset containing 27,000 64*64 Sentinel-2 patches.

\subsubsection{Implementation Details} We pre-train ResNet-18 encoders for 100 epochs using each of the four methods on each of the three pre-training datasets. Our experiments are based on Sentinel-2 only. In terms of pre-training, we use 311,667 patches for pre-training on BigEarthNet; for SEN12MS, all images are cropped to 128*128 to avoid the 50\% overlap between neighbouring patches of the original dataset; for So2Sat-LCZ42 (culture-10 version), we pre-train on the 352,366 training split. In terms of downstream evaluation, we do linear probing or fine-tuning on the training split and report testing split accuracy. 

We mainly follow the default implementation in the official repositories of the four self-supervised methods. Importantly, the queue size is 65536 and the feature dimension is 128 for MoCo-v2; the number of prototypes is 60 for SwAV; the feature dimension is 2048 for the encoder and 512 for the projector in SimSiam; three projectors with dimension 2048 are used for Barlow Twins. The data augmentations include cropping, grayscaling, Gaussian blurring, horizontal flipping and channel dropping. For simplicity, we do not include color jitterring for the main benchmark (except for SimSiam due to collapse which will be discussed later) as it is designed for natural RGB images. 4 NVIDIA A100 GPUs are used for the experiments, on which the batch size is in total 256. Links to the four methods' official sites and details of all the hyperparameters we use can be found in our code repository. In downstream tasks, we transfer 100-epoch pre-trained encoders for MoCo, SwAV and Barlow Twins, and 10-epoch pre-trained encoders for SimSiam (the model collpases afterwards). In addition, we report the performance of fully supervised learning and training with randomly initialized encoders for more thorough comparison. 
\color{black}

\subsection{Benchmark Results}

\subsubsection{General Comparsion of SSL Methods}

The general benchmark results with linear probing evaluation are shown in Fig. \ref{fig:ssl-benchmark} (a). A first conclusion is that with pure simple transfer from natural images to satellite images, all the self-supervised methods provide meaningful representations for all datasets (compared to random initialization). With frozen features that come from label-independent pre-training, we can train much fewer parameters (e.g. a full ResNet-18 has 10M parameters while the last linear layer contains only 500K) with comparable performance towards fully supervised training. 

The second observation is that MoCo-v2~\cite{chen2020improved}, which represents contrastive negative sampling, generally outperforms the other methods across datasets. This proves the robustness and transferability of using negative samples for self-supervised learning in remote sensing, which is also consistent with the fact that most of the recent self-supervised works in the remote sensing community prefer MoCo~\cite{he2020momentum} and MoCo-v2~\cite{chen2020improved} as a baseline structure.

\color{blue}
A third observation is that compared to the other methods, the training of SimSiam~\cite{chen2021exploring} is very unstable. As can be seen from the training curves in Fig. \ref{fig:ssl-benchmark} $(b)$, the training loss of SimSiam drops very quickly and significantly in the beginning epochs, and starts vibrating in the following epochs. This can be attributed to SimSiam's simplified design (as has been described in section \ref{sec:ContrastiveMethods}), which makes the similarity matching task easy to solve for the models. When utilizing checkpoints of the later epochs to serve as pre-trained weights, the model fails to beat even random initialization. That is to say, SimSiam gets collapsed after 10 epochs. Thus in real applications, it is better to add a momentum encoder or other tricks for better performance and more stable training.
\color{black}

\begin{figure}[]
    \centering
    
    \includegraphics[width=0.95\linewidth]{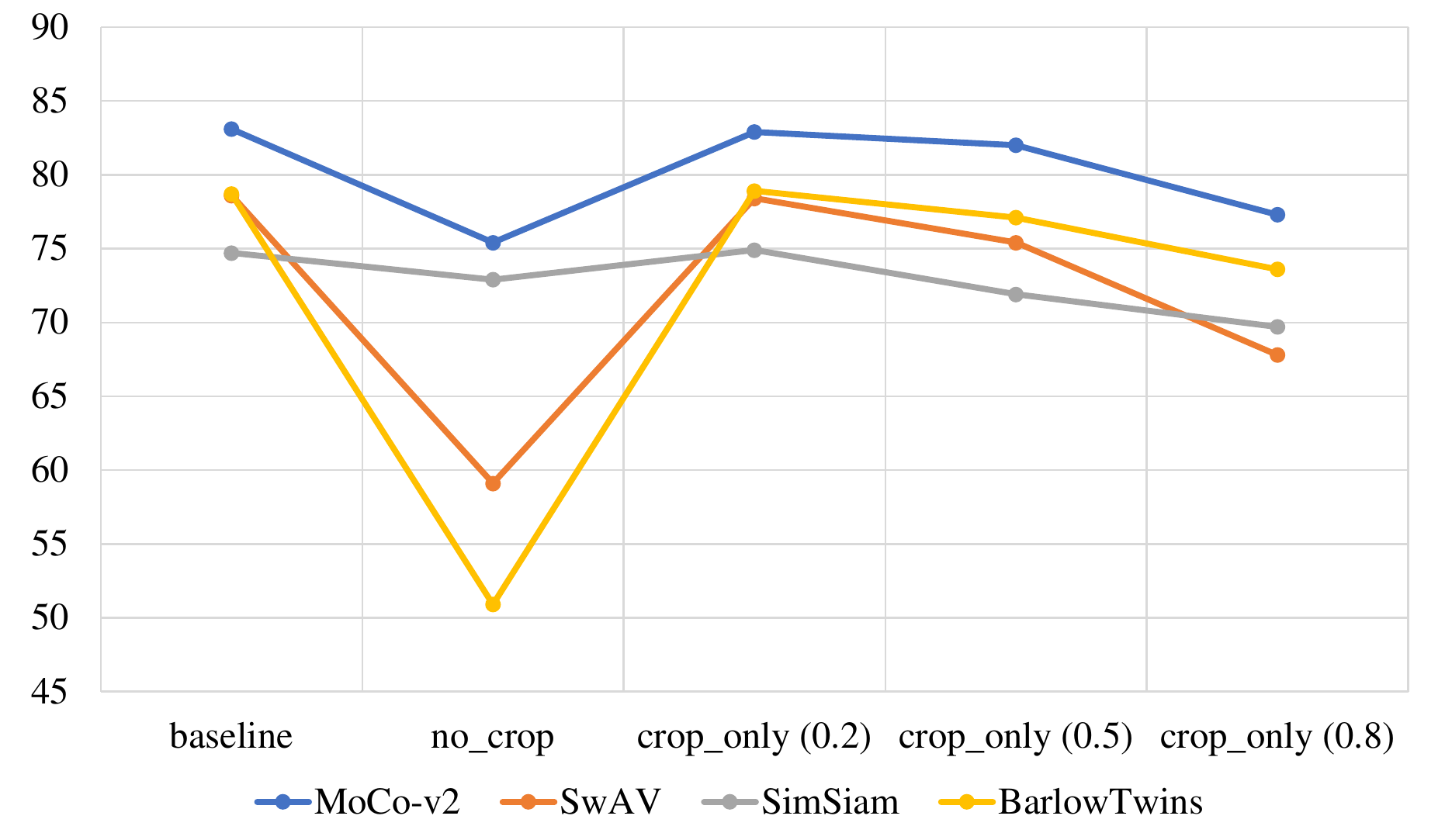}
    \caption{
    A study of data augmentation on BigEarthNet. The baseline includes all augmentations, "no\_crop" means all augmentations except cropping, and "crop\_only" means only using cropping as the augmentation with minimum cropping size $0.2$, $0.5$ and $0.8$ of the original size, respectively. 
    }
    \label{fig:rs-dataaug}
\end{figure}

\subsubsection{Data Augmentation}
In this subsection, we provide an extended study on the impact of data augmentations which have been proved very important for natural images \cite{chen2020simple}. Focusing on BigEarthNet, we do self-supervised pre-training considering a rich set of augmentations: "RandomResizedCrop", "RandomColorJitter" (a simplified version which only changes contrast and brightness), "RandomGrayscale", "RandomGaussianBlur", "RandomHorizontalFlip" and "RandomChannelDrop" (replacing one or several channels by zero). As can be seen in Fig. \ref{fig:rs-dataaug}, the results show both common and different trends compared to the computer vision community. When dropping only the "RandomResizedCrop" transform, SwAV and Barlow Twins collapse immediately and MoCo-v2 and SimSiam see a significant performance drop. All methods confirm the importance of cropping which is in line with natural images. Surprisingly, almost no visible performance drop is observed when using only cropping as the data augmentation for all four methods. This is a significant difference compared to natural images, where other augmentations (especially color distortion) bear big importance as well. Therefore, apart from cropping which is a key augmentation, further study is required on the design of other augmentations to work well on remote sensing multispectral images. In addition, to further analyze the impact of cropping, we explore different settings for the minimum cropping size. The results show that, the bigger the minimum size (less aggressive cropping), the worse the performance is. This is inline with the computer vision literature which emphasizes the importance of having stronger augmentations to make the training task hard enough for the network.

\begin{figure}
\centering
\includegraphics[width=0.87\linewidth]{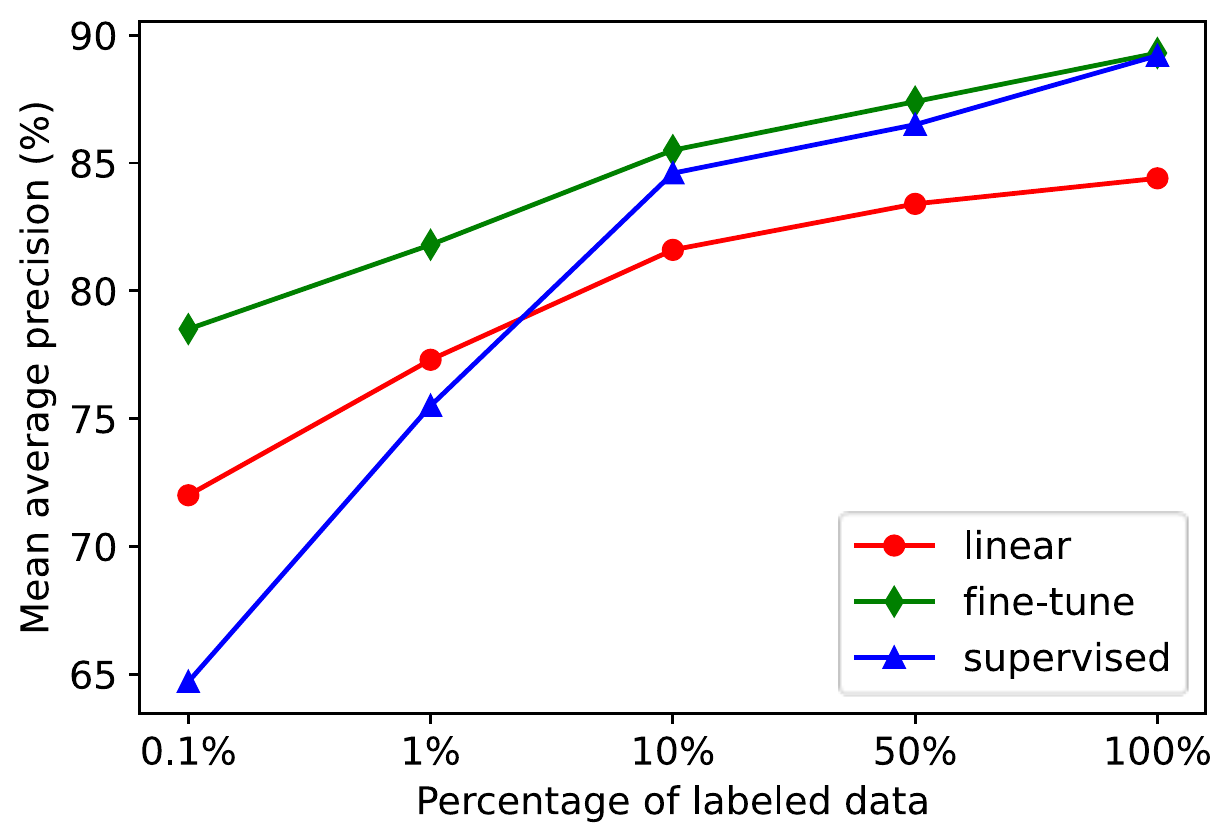}
\caption[benchmark-SSL-RS-few-shot]{
\textcolor{blue}{
BigEarthNet linear classification and fine-tuning results using different amounts of labels pre-trained with MoCo-v2. Self-supervision outperforms supervision when reducing the number of labels.}
}
\label{fig:ssl-benchmark-fewshot}
\end{figure}

\begin{figure}
\centering
\includegraphics[width=0.9\linewidth]{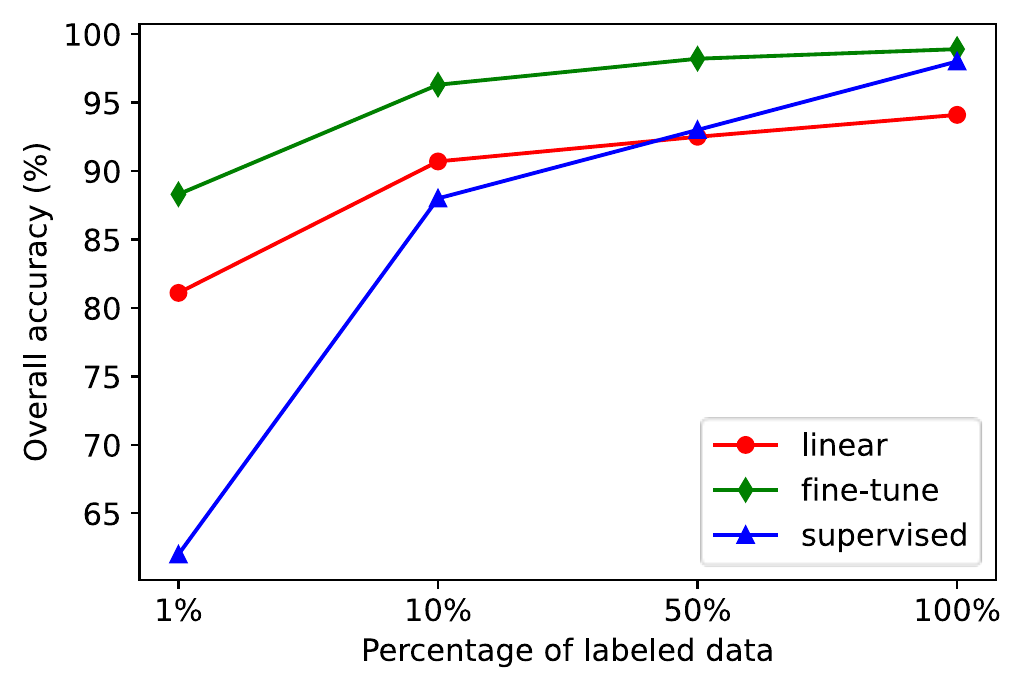}
\caption[benchmark-SSL-RS-transfer]{
\textcolor{blue}{
EuroSAT linear classification and fine-tuning results pre-trained with MoCo-v2 on BigEarthNet. Self-supervision with 50\% labels outperforms supervised learning with full labels.
}
}
\label{fig:ssl-benchmark-transfer}
\end{figure}

\subsubsection{Regime of Limited Labels}

Though Fig. \ref{fig:ssl-benchmark} proves the capability of self-supervised pre-trained models, the linear probing results are still below the upper bound of fully supervised learning with full labels. Thus, we wonder how these pre-trained models behave when reducing the number of labels and turning on fine-tuning. We pre-train ResNet18 with MoCo-v2 for 100 epochs and evaluate on BigEarthNet. Fig. \ref{fig:ssl-benchmark-fewshot} shows the linear classification and fine tuning performance when reducing the amount of labeled samples to 50\%, 10\%, 1\% and 0.1\%. The general trend of the figure proves the potential of self-supervised pre-training when we do not have enough labels: linear classification with self-supervision outperforms vanilla supervised learning when reducing the amount of available labels to 1\%. \textcolor{blue}{The advantage is more significant for fine-tuning, which outperforms supervised learning on all scenarios and provide huge improvements on tiny amount of labels. In general, the fewer labels we use, the bigger the advantage of self-supervised pre-training is.}

\subsubsection{Transfer Learning}

In addition, we report transfer learning performance by linear classification and fine-tuning on EuroSat with a ResNet18 pre-trained on BigEarthNet. As is shown in Fig. \ref{fig:ssl-benchmark-transfer}, the performance plots are similar to Fig. \ref{fig:ssl-benchmark-fewshot}, proving the pre-trained models transfer well the data representations across datasets. 
\textcolor{blue}{Promisingly, fine tuning on 50\% of labels outperforms supervised learning with full labels. In fact, fine tuning on 10\% of labels gives already very close performance compared to full-label supervised learning.}

\section{Challenges and future directions}

Self-supervised learning has been achieving great success in various vision tasks, yet it is still a relatively new branch of technologies in both computer vision and remote sensing communities. \textcolor{blue}{In this section, we summarize the challenges of self-supervised learning in remote sensing and discuss possible future directions.}

\textbf{Model collapse.} Model collapse is one of the main challenges along with the development of self-supervised learning, especially for modern contrastive learning. Though it has been shown that contrastive self-supervised learning achieved great success in the field, the fundamental theory behind those contrastive methods (especially those using no negative samples) is still not well understood. There are already some works like SimSiam~\cite{chen2021exploring} and \cite{tian2021understanding} that try to investigate the underline theory, yet they are not enough. In fact, as has been mentioned in section \ref{sec:benchmark-ssl-rs}, SimSiam easily gets collapsed in Earth observation data. Future works need to go deeper into self-supervised representation learning and explore the theoretical foundations behind model collapse in remote sensing data. 

\textbf{Pretext tasks and data augmentations.} Pretext tasks and data augmentations play a similar and very important role in self-supervised learning, as they correspond directly to what invariance the networks need to learn and what information is important for image understanding. There have been some early studies on the evaluation of different pretext tasks and data augmentations and possible improvements~\cite{mundhenk2018improvements,ryali2021leveraging}, yet they mainly focus on natural images and evaluate on image-level classification tasks. As has been proved in section \ref{sec:benchmark-ssl-rs}, the findings from common computer vision benchmarks may not completely hold for multispectral imagery, not to mention other modalities. In section \ref{sec:ssl-in-rs}, we discussed the characteristics of different remote sensing data, yet still experiments are needed. Therefore, in general more studies on pretext tasks and data augmentations are required to better understand which ones are useful for different types of remote sensing data.

\textbf{Pre-training datasets.} Most of existing remote sensing datasets are intended for supervised learning. Though they can also be used for self-supervised pre-training by discarding the labels, the amount of data is limited and the data is usually biased towards the tasks of the datasets. Therefore, it needs to be studied the necessity of self-supervised pre-training on large-scale uncurated datasets. There have been some early contributions following this thread: Leenstra \etal~\cite{leenstra2021self} proposed a Sentinel-2 multitemporal cities pairs dataset for change detection; Manas \etal~\cite{manas2021seasonal} proposed a large-scale self-supervised Sentinel-2 dataset; Heidler \etal~\cite{heidler2021self} proposed an audio-visual dataset for multi-modal self-supervised learning. However, there is still much space to explore and contribute (e.g., dataset size, image sampling strategies, and the availability of more modalities). 

\textbf{Multi-modal/temporal self-supervised learning.} As one of the most important characteristics of remote sensing data, multi-modality is a significant aspect to be explored in self-supervised representation learning \cite{chen2021self0,wang2022self,dong2020self}. On the other hand, multi-temporal image analysis is raising more interest because of the increasing frequency of data acquisition and transferring. Without the need for any human annotation, self-supervised representation learning has a big advantage to performing big data analysis which is usually a challenge for multi-modal and multi-temporal data. However, adding modalities and time stamps will also bring complexity to the model to be trained. Thus how to balance different modalities or time stamps such that the model can learn good representations from both modalities is a challenging task to be explored. 

\color{blue}
\textbf{Computing efficient self-supervised learning.} Self-supervised pre-training usually requires a large amount of computing resources because of big pre-training data, various data augmentations, the requirement of large batch sizes and more training epochs than supervised learning, etc. Few efforts have been made towards reducing computational cost, which is however an important factor for practical usage. Therefore, efficient data compression \cite{cao2021rethinking}, data loading, model design \cite{he2021masked} and hardware acceleration are necessary to be explored.
\color{black}

\textbf{Network backbones.} Most of existing self-supervised methods utilize ResNet~\cite{he2016deep} as their model backbones, while there has been a hot trend towards vision transformers (ViT)~\cite{dosovitskiy2020image}. Since ViT has shown promising results in self-supervised learning~\cite{chen2021empirical,caron2021emerging,li2021efficient}, and recent advances are exploring the link between large-scale pretraining for both language and vision~\cite{he2021masked,zhou2021ibot}, it deserves more exploration in remote sensing images as well. In addition, the specific encoder architecture itself is also important to study, as most existing methods tend to keep it unchanged with a focus on the conceptual design of self-supervision.

\textbf{Task-oriented weakly-supervised learning.} Last but not least, in parallel to the progress of task-agnostic self-supervised learning, there's also the need to integrate "few labels" or "weak labels" for better task-dependant weakly-supervised learning in practice. In other words, not only can self-supervised learning provide a pre-trained model for downstream tasks, but there's also the possibility of bringing representation learning online. For example, both data representations and the supervised task can be learned together, where the self-supervised branch helps the model capture more useful information. This scenario is especially important when moving to extreme applications that do not have a common large dataset for pre-training. In addition, this can also be a way to mitigate the generally high computational cost of training SSL algorihtms.

\section{Conclusion}

This article provides a systematic introduction and literature review of self-supervised learning for the remote sensing community. We summarize three main categories of self-supervised methods, introduce representative works and link them from natural images to remote sensing data. Moreover, we provide a preliminary benchmark and extened analysis of four modern contrastive self-supervised learning on three popular remote sensing datasets. Finally, fundamental problems and future directions are listed and discussed.

\section*{Acknowledgments}
This work was jointly supported by the European Research Council (ERC) under the European Union's Horizon 2020 research and innovation programme (grant agreement No. [ERC-2016-StG-714087], Acronym: \textit{So2Sat} and grant agreement No. [957407], Acronym: \textit{DAPHNE}), by the Helmholtz Association through the Framework of Helmholtz AI (grant number: ZT-I-PF-5-01) - Local Unit “Munich Unit @Aeronautics, Space and Transport (MASTr)”, the Helmholtz Klimainitiative (HICAM) and Helmholtz Excellent Professorship “Data Science in Earth Observation - Big Data Fusion for Urban Research” (grant number: W2-W3-100)) and by the German Federal Ministry of Education and Research (BMBF) in the framework of the international future AI lab “AI4EO – Artificial Intelligence for Earth Observation: Reasoning, Uncertainties, Ethics and Beyond” (grant number: 01DD20001). The computing resources were supported by the Helmholtz Association's Initiative and Networking Fund on the HAICORE@FZJ partition.


\small{
\printbibliography
}


\begin{IEEEbiographynophoto}{Yi Wang} received his B.E. degree in remote sensing science and technology from Wuhan University, Wuhan, China, in 2018 and his M.Sc. degree in geomatics engineering from University of Stuttgart, Stuttgart, Germany, in 2021. He is pursuing his Ph.D. degree at the German Aerospace Center, Wessling, 82234, Germany, and at Technical University of Munich, Munich, Germany. In 2020, he spent three months at perception system group, Sony Corporation, Stuttgart, Germany. His research interests include remote sensing, deep learning, computer vision and self-supervised learning. 
\end{IEEEbiographynophoto}

\begin{IEEEbiographynophoto}{Conrad M Albrecht} (M'17) received an undergraduate degree in physics from Technical University Dresden, Germany, in 2007 and a Ph.D. degree in physics with an extra certification in computer science from Heidelberg University, Germany, in 2014. Spanning the fields of physics, mathematics and computer science, among others, he was a visiting scientist with CERN, Switzerland, in 2010, and with the Dresden Max Planck Institute for the Physics of Complex Systems, Germany, in 2007. In 2015 he became research scientist with the IBM T.J. Watson Research Center, Yorktown Heights, NY, USA. Currently, since April 2021, he leads a HelmholtzAI-funded team for "Large-Scale Data Mining in Earth Observation" at the German Aerospace Center, Oberpfaffenhofen, Germany. Conrad's research agenda interconnect physical models and numerical analysis, employing Big Data technologies and machine learning. As part of the "Data Intensive Physical Analytics" team in IBM Research, he significantly contributed to industry-level solutions processing geospatial information with focus on machine-learning driven remote sensing applications. He co-organized workshops at the IEEE BigData conference and the AAAS annual meeting. Home to the US and the EU, Conrad's scientific agenda aims to strengthen transatlantic collaboration of corporate research and academia.

\end{IEEEbiographynophoto}

\begin{IEEEbiographynophoto}{Nassim Ait Ali Braham} received his M.Sc. degree in computer science from Ecole nationale Supérieure d'Informatique (ESI), Algiers, Algeria, in 2019 and M.Sc. degree in Artificial Intelligence and Data Science from Université Paris Dauphine-PSL, Paris, France, in 2020. He is pursuing his Ph.D. degree at the German Aerospace Center, Wessling, Germany, and at Technical University of Munich, Munich, Germany. In 2019, he spent six months at the LIRIS-CNRS laboratory, Lyon, France. In 2020, he spent six months at the LAMSADE-CNRS laboratory, PSL Research University, Paris, France. His research interests include deep learning, computer vision, self-supervised learning and remote sensing.
\end{IEEEbiographynophoto}

\begin{IEEEbiographynophoto}{Lichao Mou} received the Bachelor's degree in automation from the Xi'an University of Posts and Telecommunications, Xi'an, China, in 2012, the Master's degree in signal and information processing from the University of Chinese Academy of Sciences (UCAS), China, in 2015, and the Dr.-Ing. degree from the Technical University of Munich (TUM), Munich, Germany, in 2020. He is currently a Guest Professor at the Munich AI Future Lab AI4EO, TUM and the Head of Visual Learning and Reasoning team at the Department ``EO Data Science'', Remote Sensing Technology Institute (IMF), German Aerospace Center (DLR), Wessling, Germany. Since 2019, he is a Research Scientist at DLR-IMF and an AI Consultant for the Helmholtz Artificial Intelligence Cooperation Unit (HAICU). In 2015 he spent six months at the Computer Vision Group at the University of Freiburg in Germany. In 2019 he was a Visiting Researcher with the Cambridge Image Analysis Group (CIA), University of Cambridge, UK. He was the recipient of the first place in the 2016 IEEE GRSS Data Fusion Contest and finalists for the Best Student Paper Award at the 2017 Joint Urban Remote Sensing Event and 2019 Joint Urban Remote Sensing Event.
\end{IEEEbiographynophoto}

\begin{IEEEbiographynophoto}{Xiao Xiang Zhu} (S'10--M'12--SM'14--F'21) received the Master (M.Sc.) degree, her doctor of engineering (Dr.-Ing.) degree and her “Habilitation” in the field of signal processing from Technical University of Munich (TUM), Munich, Germany, in 2008, 2011 and 2013, respectively.
\par
She is the Chair Professor for Data Science in Earth Observation at Technical University of Munich (TUM) and the Head of the Department ``EO Data Science'' at the Remote Sensing Technology Institute, German Aerospace Center (DLR). Since 2019, Zhu is a co-coordinator of the Munich Data Science Research School (www.mu-ds.de). Since 2019 She also heads the Helmholtz Artificial Intelligence -- Research Field ``Aeronautics, Space and Transport". Since May 2020, she is the PI and director of the international future AI lab "AI4EO -- Artificial Intelligence for Earth Observation: Reasoning, Uncertainties, Ethics and Beyond", Munich, Germany. Since October 2020, she also serves as a co-director of the Munich Data Science Institute (MDSI), TUM. Prof. Zhu was a guest scientist or visiting professor at the Italian National Research Council (CNR-IREA), Naples, Italy, Fudan University, Shanghai, China, the University  of Tokyo, Tokyo, Japan and University of California, Los Angeles, United States in 2009, 2014, 2015 and 2016, respectively. She is currently a visiting AI professor at ESA's Phi-lab. Her main research interests are remote sensing and Earth observation, signal processing, machine learning and data science, with their applications in tackling societal grand challenges, e.g. Global Urbanization, UN’s SDGs and Climate Change.

Dr. Zhu is a member of young academy (Junge Akademie/Junges Kolleg) at the Berlin-Brandenburg Academy of Sciences and Humanities and the German National  Academy of Sciences Leopoldina and the Bavarian Academy of Sciences and Humanities. She serves in the scientific advisory board in several research organizations, among others the German Research Center for Geosciences (GFZ) and Potsdam Institute for Climate Impact Research (PIK). She is an associate Editor of IEEE Transactions on Geoscience and Remote Sensing and serves as the area editor responsible for special issues of IEEE Signal Processing Magazine. She is a Fellow of IEEE.
\end{IEEEbiographynophoto}

\end{document}